\documentclass[10pt,twocolumn,letterpaper]{article}

\usepackage[pagenumbers]{cvpr} %

\usepackage{graphicx}
\usepackage{amsmath}
\usepackage{amssymb}
\usepackage{booktabs}

\usepackage[pagebackref,breaklinks,colorlinks]{hyperref}

\usepackage[capitalize]{cleveref}
\crefname{section}{Sec.}{Secs.}
\Crefname{section}{Section}{Sections}
\Crefname{table}{Table}{Tables}
\crefname{table}{Tab.}{Tabs.}

\def\cvprPaperID{6623} %
\def\confName{CVPR}
\def\confYear{2022}

\usepackage{overpic}
\usepackage{enumitem} %
\usepackage{overpic} %
\usepackage{color}
\usepackage{mathtools} %
\usepackage{multirow}
\usepackage{graphicx,booktabs,array}
\usepackage{colortbl}
\usepackage[dvipsnames]{xcolor}
\usepackage{microtype}
\usepackage{lscape}
\usepackage{rotating}

\definecolor{turquoise}{cmyk}{0.65,0,0.1,0.3}
\definecolor{purple}{rgb}{0.65,0,0.65}
\definecolor{dark_green}{rgb}{0, 0.5, 0}
\definecolor{orange}{rgb}{0.8, 0.6, 0.2}
\definecolor{red}{rgb}{0.8, 0.2, 0.2}
\definecolor{darkred}{rgb}{0.6, 0.1, 0.05}
\definecolor{blueish}{rgb}{0.0, 0.3, .6}
\definecolor{light_gray}{rgb}{0.7, 0.7, .7}
\definecolor{pink}{rgb}{1, 0, 1}
\definecolor{greyblue}{rgb}{0.25, 0.25, 1}

\definecolor{mydarkblue}{rgb}{0,0.08,0.55}
\hypersetup{  %
    colorlinks=true,
    linkcolor=mydarkblue,
    citecolor=mydarkblue,
    filecolor=mydarkblue,
    urlcolor=mydarkblue
}

\newcommand{\todo}[1]{{\color{red}#1}}
\newcommand{\TODO}[1]{\textbf{\color{red}[TODO: #1]}}

\newcommand{\AT}[1]{{\color{blueish}{\bf [AT: #1]}}} %
\newcommand{\At}[1]{\marginpar{\tiny{\textcolor{blueish}{#1}}}} %
\newcommand{\dd}[1]{{\color{dark_green}#1}}
\newcommand{\Dd}[1]{\marginpar{\tiny{\textcolor{dark_green}{#1}}}}
\newcommand{\DD}[1]{{\color{dark_green}{\bf [DD: #1]}}}
\newcommand{\SV}[1]{{\color{orange}{\bf [SV: #1]}}} %
\newcommand{\NR}[1]{{\color{red}{\bf [NR: #1]}}} %
\newcommand{\TF}[1]{{\color{purple}{\bf [TF: #1]}}} %

\renewcommand{\todo}[1]{}
\renewcommand{\TODO}[1]{}

\renewcommand{\AT}[1]{}
\renewcommand{\At}[1]{}
\renewcommand{\dd}[1]{#1}
\renewcommand{\Dd}[1]{}
\renewcommand{\DD}[1]{}
\renewcommand{\SV}[1]{} %
\renewcommand{\NR}[1]{} %
\renewcommand{\TF}[1]{} %

\newcommand{\CIRCLE}[1]{\raisebox{.5pt}{\footnotesize \textcircled{\raisebox{-.6pt}{#1}}}}

\newcommand\boldpnov[1]{\noindent\textbf{#1}\hspace{.2cm}}
\usepackage{array}
\newcolumntype{P}[1]{>{\centering\arraybackslash}p{#1}}
\newcommand\CoAuthorMark{\footnotemark[\arabic{footnote}]}

\newcommand{\loss}[1]{\mathcal{L}_\text{#1}}
\newcommand{\expect}{\mathbb{E}}

\newcommand{\Figure}[1]{Figure~\ref{fig:#1}}

\newcommand{\Table}[1]{Table~\ref{tab:#1}}

\newcommand{\Section}[1]{Section~\ref{sec:#1}}

\usepackage{blindtext}

\usepackage{lipsum}

\renewcommand{\paragraph}[1]{\vspace{.25em}\noindent\textbf{#1}.}
\newcommand{\bigname}[1]{\vspace{.2em}\noindent\textbf{#1}}

\newcommand{\NeSF}{NeSF\xspace}
\newcommand{\NeRF}{NeRF\xspace}
\newcommand{\MultiShapeNet}{\texttt{ToyBox}\xspace}

\newcommand{\ToyBoxEasy}{\texttt{ToyBox5}\xspace}
\newcommand{\ToyBoxHard}{\texttt{ToyBox13}\xspace}
\newcommand{\KLEVR}{\texttt{KLEVR}\xspace}
\newcommand{\KLEVRShort}{\texttt{KLEVR}\xspace}
\newcommand{\mIoU}{mIoU\xspace}
\newcommand{\DeepLab}{DeepLab\xspace}
\newcommand{\SparseConvNet}{SparseConvNet\xspace}

\newcommand{\SupplementaryMaterial}{{\color{purple}supplementary material}\xspace}
\newcommand{\NumTrainImagesPerScene}{9\xspace}
\newcommand{\NumTestImagesPerScene}{4\xspace}
\newcommand{\NumRgbImagesPerScene}{210\xspace}
\newcommand{\NumKLEVRTrainScenes}{100\xspace}
\newcommand{\NumToyBoxTrainScenes}{500\xspace}
\newcommand{\NumKLEVRTestScenes}{20\xspace}
\newcommand{\NumToyBoxTestScenes}{25\xspace}
\newcommand{\TrainViews}{\textsc{train~cameras}\xspace}
\newcommand{\TestViews}{\textsc{test~cameras}\xspace}
\newcommand{\TrainScenes}{\textsc{train~scenes}\xspace}
\newcommand{\TestScenes}{\textsc{novel~scenes}\xspace}
\newcommand{\WatchTheVideo}{\textbf{We strongly encourage the reader to view additional results in the \href{\YouTubeLink}{accompanying video} for more detail.\xspace}}

\newcommand{\ProjectWebsiteLink}{http://example.com}
\newcommand{\YouTubeLink}{NOT_DEFINED}
\newcommand{\OverviewVideoImage}{NOT_DEFINED}

\renewcommand{\YouTubeLink}{https://www.youtube.com/watch?v=yagzIwmiu20}
\renewcommand{\OverviewVideoImage}{fig/overview_video/video_arxiv}
\renewcommand{\ProjectWebsiteLink}{https://nesf3d.github.io}

\begin{document}
\title{NeSF: Neural Semantic Fields \\ for Generalizable Semantic Segmentation of 3D Scenes}

\newcommand{\google}{1}
\newcommand{\uoft}{2}
\author{
Suhani Vora\textsuperscript{\google}\thanks{Denotes equal contribution.} \and
Noha Radwan\textsuperscript{\google}\CoAuthorMark{} \and
Klaus Greff\textsuperscript{\google} \and
Henning Meyer\textsuperscript{\google} \and
Kyle Genova\textsuperscript{\google} \and
Mehdi S. M. Sajjadi\textsuperscript{\google} \and
Etienne Pot\textsuperscript{\google} \and
Andrea Tagliasacchi\textsuperscript{\google,\uoft} \and
Daniel Duckworth\textsuperscript{\google} \and
\\
\textsuperscript{\google}Google Research $\quad$
\textsuperscript{\uoft}University of Toronto $\quad$
}
\maketitle
\begin{abstract}
We present \emph{\NeSF{}}, a method for producing 3D semantic fields from posed RGB images alone.
In place of classical 3D representations, our method builds on recent work in implicit neural scene representations wherein 3D structure is captured by point-wise functions.
We leverage this methodology to recover 3D density fields upon which we then train a 3D semantic segmentation model supervised by posed 2D semantic maps.
Despite being trained on 2D signals alone, our method is able to generate 3D-consistent semantic maps from novel camera poses and can be queried at arbitrary 3D points.
Notably, \NeSF{} is compatible with any method producing a density field, and its accuracy improves as the quality of the density field improves.
Our empirical analysis demonstrates comparable quality to competitive 2D and 3D semantic segmentation baselines on complex, realistically-rendered synthetic scenes.
Our method is the first to offer truly dense 3D scene segmentations requiring only 2D supervision for training, and does not require any semantic input for inference on novel scenes.
We encourage the readers to visit the \href{\ProjectWebsiteLink}{project website}.
\end{abstract}

\section{Introduction}
\label{sec:intro}
High-level semantic understanding of 3D scenes as captured by digital images and videos is a fundamental objective of computer vision.
Well-studied tasks such as scene classification~\cite{krizhevsky2012imagenet}, object detection~\cite{liu2020deep}, semantic segmentation~\cite{lateef2019survey}, and instance segmentation~\cite{hafiz2020survey} infer semantic descriptions of scenes from RGB and other sensors and form the foundation for applications such as visual navigation~\cite{bonin2008visual} and robotic interaction~\cite{billard2019trends}.

The most common approach to scene understanding is to narrow the scope to 2D~(image-space) reasoning, wherein classical image-to-image architectures~\cite{ronneberger2015unet} are trained on large collections of semantically-annotated images~\cite{lin2014mscoco}. 
These methods, while straightforward to apply, only produce per-pixel annotations and largely ignore the underlying 3D structure of the scene.   
Instead, our goal is to use a set of RGB images with known poses to produce a 3D \textit{semantic field}:~a function mapping 3D positions to probability distributions over semantic categories.

% Figure from this slide,
% https://docs.google.com/presentation/d/1LFiVMtnXfRPOX8m_1mzIiozypevWvzJjGnjURadDinc/edit?resourcekey=0-0YUVXmQENSR0hI-G7BIYhA#slide=id.gfb9da2db54_0_131

\begin{figure}[t]
\centering
    \includegraphics[width=\linewidth]{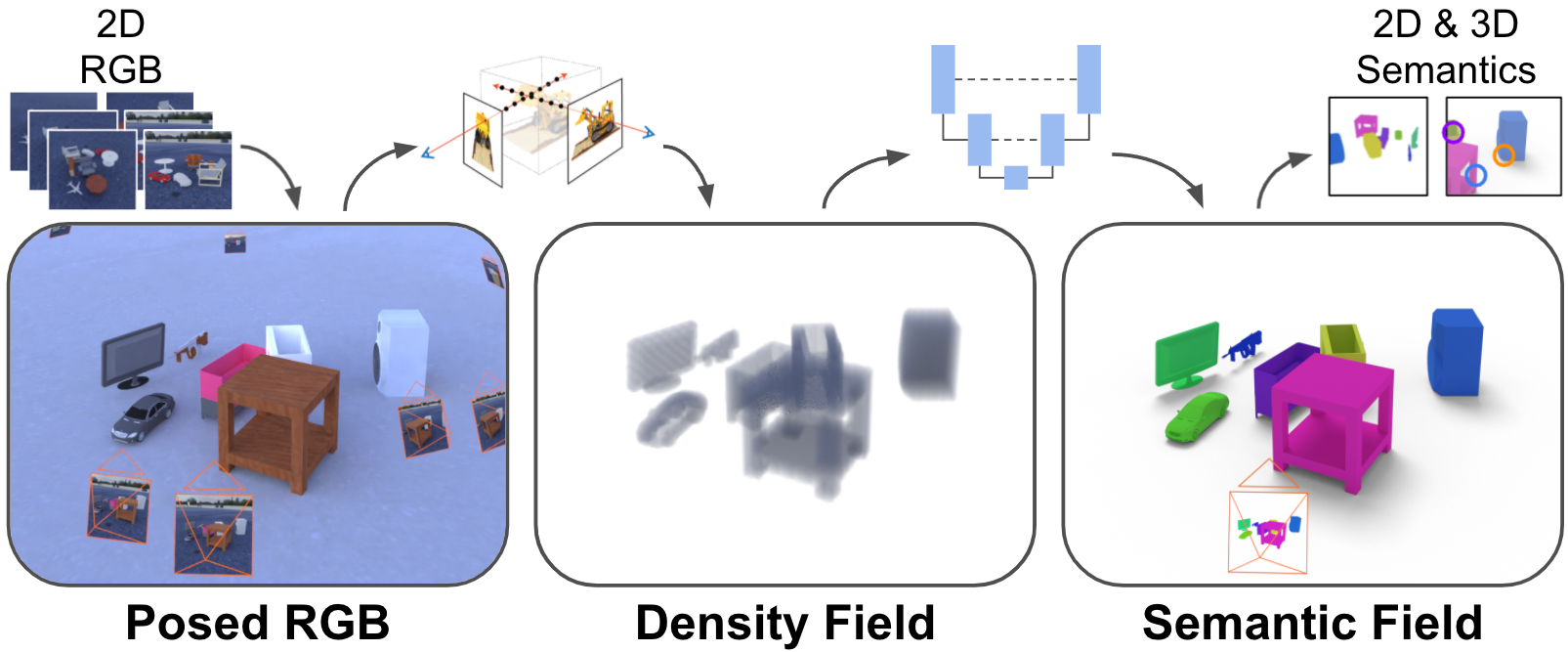}
\vspace{-1em}
\caption{
    \textbf{Overview -- }
    We train our method on collections of posed 2D RGB images and 2D semantic maps, each collection describing an independent scene.
    Given a new set of posed 2D RGB images, we extract an implicit volumetric representation of the scene's 3D geometry and infer a 3D semantic field.
    The semantic field can then be used to render dense 2D semantic maps from novel camera poses or queried directly in 3D.
    Our method generalizes to novel scenes, and requires as little as one semantic map per scene at training time.
    We encourage readers to visit the \href{\ProjectWebsiteLink}{project website}.
}
\label{fig:teaser}
\end{figure}

For 3D semantic segmentation, most prior work relies on 3D sensors~\cite{izadi2011kinectfusion, douillard2011segmentation, kowdle2018need4speed} and/or 3D semantic labels~\cite{dai2017scannet, caesar2019nuscenes}.  
Convolutional network architectures have been designed for 3D point clouds~\cite{qi2017pointnet,choy2019minkowskinet}, voxel grids~\cite{tran20153dconvnets}, and polygonal meshes~\cite{hanocka2019meshcnn,masci2015geodesic}. 
However, 3D sensors are not as affordable nor as widely available as their RGB cameras, and 3D annotations are significantly more challenging to produce than their 2D counterparts and are hence generally scarce~\cite{genova2021learning}.

To overcome this challenge, researchers have adopted \textit{hybrid} 2D/3D reasoning, so as to propagate densely supervised semantic signals from 2D projections back to an underlying 3D substrate~\cite{dai20183dmv,genova2021learning,kundu2020virtual}.
At test time, these methods still require a classical 3D representation to be provided as input, hence limiting their applicability and performance.
One interesting exception is Atlas~\cite{Murez20eccv_atlas}, which only requires posed photos at test time but still requires 3D supervision to train.

In parallel to these developments, a new family of 3D representations have emerged based on implicit, coordinate-wise functions~\cite{chen2019imnet,park2019deepsdf,mescheder2019occupancy}.
In this regime, one trains a neural network to predict quantities such as occupancy, signed distance, density and radiance.
Unlike explicit representations of geometry, neural implicit methods are memory efficient and are able to capture impressive levels of geometric detail~\cite{Takikawa21cvpr_nglod}.
However, the vast majority of methods built on this approach target computer graphics applications such as novel view synthesis~\cite{Wang21cvpr_IBRNet,zhang2021nerfactor,Liu20neurips_sparse_nerf,Rebain20arxiv_derf,Yu21cvpr_pixelNeRF} without an interpretable semantic understanding of the scene.
One notable exception is Semantic-NeRF~\cite{Zhi21iccv_SemanticNeRF}, which regresses a per-3D point semantic class in addition to radiance and density.
Similar to \NeRF{}~\cite{Mildenhall20eccv_nerf}, this method is only applicable to novel \emph{views} within the same scene and does not provide the form of generalization one expects in classical semantic segmentation: the ability to infer semantics on novel \emph{scenes}.

We introduce~\emph{Neural Semantic Fields~(\NeSF{})}, a method for semantic segmentation of 3D scenes via image-space semantic annotations; see~\Figure{teaser}.
Unlike Semantic-\NeRF{}, \NeSF{} generalizes to scenes unobserved at training time.
In place of an explicit scene representation, \NeSF{} builds on the implicit representations of geometry recovered by methods such as \NeRF{}. 
In particular, we apply a neural network to \NeRF{}'s density field to obtain what we refer to as a scene's \emph{semantic field}. 
A semantic field is thus defined as a coordinate-wise function mapping 3D points to probability distributions over semantic categories.
Similar to Semantic-\NeRF{}, we apply the volumetric rendering equation to generate 2D semantic maps, enabling supervision from posed semantic annotations in image-space.
To the best of our knowledge, \NeSF{} is the first method capable of producing dense 2D and 3D segmentations of novel scenes from posed RGB images alone.
The ability to reason about 3D information from 2D supervision alone is essential to the deployment of 3D computer vision at scale; while 2D sensors are ubiquitous, 3D sensors are expensive, unwieldy, and unlikely to be deployed in the mass market.

As large scale datasets of 3D semantically annotated scenes are scarce,
we propose three novel datasets of increasing complexity: \KLEVR{}, \ToyBoxEasy{}, and \ToyBoxHard{}.
While datasets for 2D and 3D semantic scene understanding already exist~\cite{dai2017scannet, chang2017matterport3d, straub2019replica, caesar2020nuscenes}, they lack the scale, detail, realism, and precision necessary to simultaneously evaluate 2D and 3D semantic segmentation. 
We construct over 1,000 scenes of randomly-placed, toy-sized objects and render hundreds of RGB images with realistic lighting and materials of each. Notably, random object placement breaks relational consistencies between objects that exist in available datasets, enabling harder tasks. Each RGB image is paired with corresponding ground-truth camera intrinsics and extrinsics, a semantic map and a depth map. We evaluate our method on these three datasets and compare its performance to competitive techniques in 2D and 3D scene understanding.

\paragraph{Contributions}
\vspace{-.5em}
\begin{itemize}[leftmargin=*]
\setlength\itemsep{-.3em}
\item We introduce the first method for generating 3D semantic fields for novel scenes trained solely on posed RGB images and semantic maps. Unlike prior work, our method (i) can be queried anywhere within a bounded 3D volume, (ii) is capable of rendering semantic maps from novel camera poses, and (iii) generalizes to novel scenes with as few as one semantic map per scene at training time.
\item We propose three new synthetic datasets for 2D and 3D semantic scene understanding.
In total, these datasets comprise of over 1,000 scenes and 3,000,000 realistically-rendered and semantically annotated frames.
Upon publication, these datasets will be released to the community alongside code to reproduce them.
\end{itemize}

\section{Related works}
\label{sec:related}
We now briefly overview related work in semantic segmentation~\cite{minaee2021image, he2021deep} and 3D reconstruction~\cite{jin20203d}.

\paragraph{Semantic segmentation} 
Semantic segmentation is a heavily researched area, with most methods targeting a fully supervised, single modality problem (2D~\cite{chollet2017xception, chen2017rethinking, long2015fully, ronneberger2015u, chen2017deeplab, badrinarayanan2017segnet} or 3D~\cite{zhu2020cylindrical, cheng20212, che2019object, guo2020deep, liu2019deep, cortinhal2020salsanext, milioto2019iros, zhang2020polarnet}). 
2D approaches like \DeepLab{}~\cite{chen2017deeplab} train a CNN to segment each pixel in an image. 
There are also analogous approaches in 3D for various shape representations -- point clouds~\cite{wang2019towards, pham2019jsis3d,qi2017pointnet,qi2017pointnet++,shi2019pv,thomas2019kpconv,jsenet}, sparse or dense voxel grids~\cite{song2017semantic, dai20183dmv,choy20194d,choy2019fully,graham20183d,han2020occuseg, riegler2017octnet}, or meshes~\cite{hanocka2019meshcnn,huang2019texturenet}. 
In contrast to these methods, our method reconstructs and then segments a dense 3D representation from 2D inputs and supervision alone, and does not require ground truth 3D annotations or input geometry.

\paragraph{Hybrid and multi-modal methods} Many methods use one data modality to supervise or inform another~\cite{owens2016ambient, zhao2018soundofpixels, korbar2018cooperative, gan2019self, aytar2016soundnet, ephrat2018looking, koepke2020sight, alwassel2020self, nagrani2020speech2action, gupta2016cross, meyerimproving, girdhar2019distinit, tian2019contrastive, jing2020self, lawin2017deep}. 
For 3D semantic segmentation, multiview fusion~\cite{hermans2014dense,kundu2020virtual,vineet2015incremental, armeni_iccv19, mccormac2017semanticfusion, mascaro2021diffuser, ma2017multi,vineet2015incremental,zhang2019large} is a popular family of methods that require only image supervision. 
However, these methods reason exclusively in the image domain and require an input 3D substrate such as a point cloud or polygonal mesh on which to aggregate 2D information. 
Similarly, Genova \etal \cite{genova2021learning} propose a method for 3D point cloud segmentation from 2D supervision, but still requires input 3D geometry.
In a separate line of work, researchers have proposed pipelines for 3D segmentation that benefit from 2D image features~\cite{kundu2020virtual, jaritz2019multi, dai20183dmv, 6907298, vora2020pointpainting}.
Unlike our method, these approaches also require a full 3D supervision.

\paragraph{Implicit representations}
Most similar to our approach, Atlas~\cite{Murez20eccv_atlas} learns a 3D implicit TSDF reconstruction from 2D images while also learning to segment the predicted scene geometry.
However, this approach requires ground truth 3D data and supervision, while our method requires only images at \emph{both} train and test time.
Other methods use implicit representations to reconstruct a 3D scene~\cite{Sucar21iccv_iMAP, Peng2020ECCV} or shape~\cite{Chen20cvpr_bsp, Chibane20cvpr_ifnet, Deng20cvpr_cvxnet, Mescheder19cvpr_occupancy_net, Chen19cvpr_IM_NET, Niemeyer20cvpr_DVR,Peng21cvpr_NeuralBody, Genova19iccv_sif, Oechsle21iccv_UNISURF, Park19cvpr_deepsdf, Takikawa21cvpr_nglod, Chabra20eccv_DLS, Saito19Iccv_PIFu, Deng20eccv_NASA}.
A more recent work approaches the problem with image supervision only~\cite{Sucar21iccv_iMAP} but does not consider semantics.

\begin{figure*}
\begin{center}
\includegraphics[width=.99\linewidth]{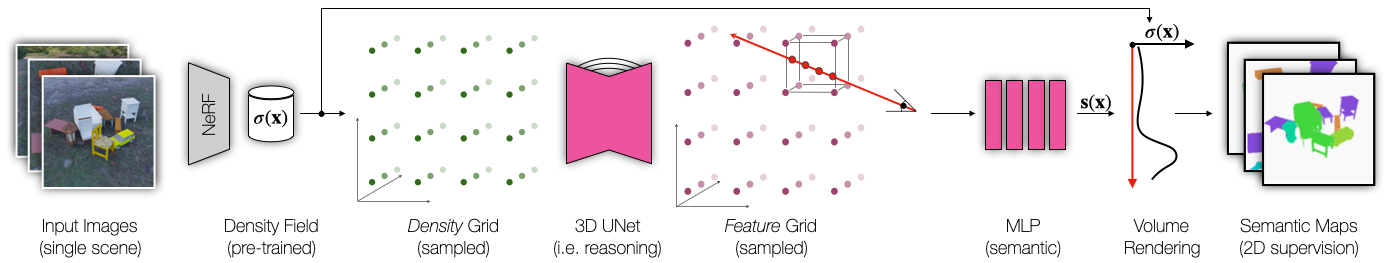}
\end{center}
\vspace{-1em}
\caption{
\textbf{Architecture -- }
Given a pre-trained NeRF model, we sample its volumetric density grid to obtain the 3D scene representation. This grid is converted to a semantic-feature grid by employing a fully convolutional volume-to-volume network thus allowing for geometric reasoning. The semantic-feature grid is in turn translated to semantic probability distributions using the volumetric rendering equation. Note the semantic 3D UNet is trained across all scenes in the \TrainScenes set, though not explicitly depicted for the sake of simplicity. Additionally, note that NeSF is trained \textit{solely} using 2D supervisory signals and that no segmentation maps are provided at test time.
% Our technique receives a pre-trained NeRF model as input, sample its volumetric density, employs a fully convolutional volume-to-volume layer to allow geometric reasoning to take place, and then translates the feature grid into the semantic class along a ray by volume rendering; the entire training process is \textit{solely} supervised by 2D segmentation maps, and no segmentation maps are provided at test time.
}
\label{fig:overview}
\end{figure*}

\paragraph{Neural radiance fields} Recently, a variety of methods based on \NeRF{}~\cite{Mildenhall20eccv_nerf} have become popular for novel view synthesis~\cite{Lindell20cvpr_AutoInt, MartinBrualla21cvpr_nerfw, Srinivasan21cvpr_NeRV, Neff21egsr_DONeRF, Barron21iccv_Mip_NeRF, Garbin21iccv_FastNeRF, Hedman21iccv_baking_NeRF, Jeong21iccv_SCNeRF, Lin21iccv_BARF, Park21iccv_nerfies, Reiser21iccv_KiloNeRF, Yu21iccv_PlenOctrees, Liu20neurips_sparse_nerf, Wang21cvpr_IBRNet}, 3D reconstruction~\cite{Rematas21icml_ShaRF, Yariv20neurips_MVNeuralSurfaceRecon, Tretschk21iccv_NR_NeRF, Jain21iccv_DietNeRF, Noguchi21iccv_NARF, Jang21iccv_CodeNeRF, Henzler21cvpr_object_categories, DeVries21iccv_GSN, Boss21iccv_NeRD, Yu21cvpr_pixelNeRF, Yuan21cvpr_STaR, Boss21iccv_NeRD, Chen21iccv_MVSNeRF, Cole21iccv_RtS}, generative modeling~\cite{Meng21iccv_GNeRF, Schwarz20neurips_graf, Kosiorek21icml_NeRF_VAE, Niemeyer21cvpr_GIRAFFE} and semantic segmentation~\cite{Zhi21iccv_SemanticNeRF}. 
The majority of these models demonstrate impressive results on novel view synthesis but are only applicable in the single-scene setting.
Others generalize to novel scenes but focus on novel view synthesis or reconstruction.
In contrast, ours is the first approach capable of generating 3D semantic segmentations of novel scenes without supervision at test time.
\newcommand{\semantic}{\mathcal{S}}
\newcommand{\semfield}{\mathbf{s}}
\newcommand{\C}{\mathcal{C}} %
\newcommand{\image}{\C}
\newcommand{\gt}{\text{gt}} %
\newcommand{\params}{\boldsymbol{\theta}}
\newcommand{\near}{{t_n}}
\newcommand{\far}{{t_f}}
\newcommand{\radiance}{\mathbf{c}}
\newcommand{\sky}{\text{sky}}
\newcommand{\ray}{\mathbf{r}}
\newcommand{\Rays}{\mathbf{\mathcal{R}}}
\newcommand{\point}{\mathbf{x}}
\newcommand{\origin}{\mathbf{o}}
\newcommand{\dir}{\mathbf{d}}
\newcommand{\density}{\boldsymbol{\sigma}}
\newcommand{\feature}{\mathbf{z}}
\newcommand{\definedAs}{\coloneqq}
\newcommand{\lidarPoints}{\mathcal{D}}
\newcommand{\scene}{\mathcal{S}}
\newcommand{\iScene}{s}
\newcommand{\nScenes}{S}
\newcommand{\iCamera}{c}
\newcommand{\nCameras}{C}
\newcommand{\camera}{\mathbf{\gamma}}
\newcommand{\given}{{\,|\,}}
\newcommand{\translator}{\mathcal{T}}
\newcommand{\densityGrid}{\Sigma}
\newcommand{\latticeResolution}{\epsilon}
\newcommand{\featureGrid}{\mathcal{F}}
\newcommand{\RealNumbers}{\mathbb{R}}
\newcommand{\Integers}{\mathbb{Z}_{+}}
\newcommand{\NumCameraParams}{\Gamma}
\newcommand{\NesfParams}{\boldsymbol{\tau}}
\newcommand{\unet}{\text{unet}}
\newcommand{\mlp}{\text{mlp}}
\newcommand{\TriLerp}{\text{TriLerp}}
\newcommand{\Angle}{\gamma}
\newcommand{\RandRot}{R}

\section{Method}
We train \NeSF{} on a collection of $\nScenes$ scenes, each described by a collection of RGB images~$\{\image^\gt_{\iScene,\iCamera} \in [0, 1]^{H \times W \times 3}\}$ and paired to a collection of semantic maps~$\{\semantic^\gt_{\iScene,\iCamera} \in \Integers^{H \times W}\}$.
Both images and maps are indexed by camera index~$\iCamera$ and a scene index~$\iScene$.
For the sake of exposition, we assume that each RGB map is paired with a semantic map and that each scene contains $\nCameras$ map pairs, but the method itself makes no such assumption.
Similar to prior work, we also assume the availability of camera calibration parameters~$\{\camera_{\iScene,\iCamera} \in \RealNumbers^{\NumCameraParams}\}$ providing an explicit connection between each pixel and the 3D ray $\ray$ cast within the 3D scene. We consider the problem of jointly estimating the camera calibration and the scene representation outside the context of this work, see~\cite{Wang21arxiv_NeRFminusminus, Lin21iccv_BARF, YenChen20iros_iNeRF}.
Our method involves two stages, which are described in the following subsections:
\vspace{-1em}
\begin{itemize}[leftmargin=*]
\setlength\itemsep{-.3em}
\item{
    \textbf{\Section{preliminaries}}: In the first stage, we pre-train neural radiance fields on posed RGB maps $\{(\image^\gt_{\iScene,\iCamera},  \camera_{\iScene,\iCamera})\}$ \textit{independently} for each scene $\iScene \in [1 \cdots \nScenes]$.
    This results in a set of neural radiance fields with network parameters $\{\theta_\iScene\}$. 
    To focus on the core task of understanding from 3D geometry, we disregard the radiance portion of these fields and employ the \textit{volumetric density fields} $\density(\point \given \theta_\iScene) \in [0, \infty)$ below. 
}
\item{
    \textbf{\Section{semantic}}: 
    In the second stage, we train a density-to-semantics \textit{translation} network $\translator$ parameterized by $\NesfParams =  \{ \NesfParams_{\unet}, \NesfParams_{\mlp} \}$.
    Given a scene's 3D geometry represented by density field~$\density_{\iScene} = \density(\cdot \given \theta_\iScene)$, this network produces a 3D semantic field~$\semfield(\point \given \density_\iScene, \NesfParams)$ assigning each point a probability distribution over semantic categories.
    While the translation network produces a 3D field, we apply the volumetric rendering equation to obtain 2D semantic maps from reference camera poses  $\{ \camera_{\iScene,\iCamera} \}$.
    Predicted semantic maps can then be compared to their ground truth counterparts $\{ \semantic^\gt_{\iScene,\iCamera} \}$ in a differentiable way.
}
\end{itemize}

\subsection{\NeRF pre-training}
\label{sec:preliminaries}
To extract an accurate, dense representation of each scene's 3D geometry, we leverage neural radiance fields as proposed in~\cite{Mildenhall20eccv_nerf}.
To simplify notation, we drop scene index $\iScene$ for the remainder of this section as all scenes can be trained independently and in parallel.
More specifically, given a collection of posed RGB images~$\{ (\image_\iCamera^\gt, \camera_\iCamera) \}$, and denoting with $\ray \sim \Rays(\camera_\iCamera)$ rays corresponding to pixels from image $\image_\iCamera^\gt$, a neural radiance field model with parameters $\theta$ is trained by minimizing the squared photometric reconstruction loss:
\begin{equation}
\loss{rgb}(\params) = \sum_\iCamera \expect_{\ray \sim \Rays(\camera_\iCamera)}
\left[
||\C(\ray \given \theta) - \C^\gt_\iCamera(\ray) ||_2^2 
\right]
\label{eq:rgb}
\end{equation}

\vspace{-.5em}
\noindent
where $\C^\gt_\iCamera(\ray)$ is the ground truth color of ray passing through a pixel in image $\iCamera$, and the color $\C(\ray \given \theta)$ is computed by applying the volumetric rendering equation with the ray's near and far bounds $t \in [\near, \far]$:
\begin{align}
\C(\ray \given \theta) = \int_\near^\far 
w(t \given \theta)
\cdot
\radiance(t \given \theta) \, dt
\label{eq:color}
\end{align}

\vspace{-.5em}
\noindent
Let $\ray(t) {=} \origin + t\mathbf{d}$ represent a point along a ray with origin $\origin$ and direction $\mathbf{d}$. Weight $w(t){=}w(\ray(t))$ is then defined as:
\begin{align}
w(t) = \underbrace{\exp \left( -\int_{t_n}^{t} \density(u) \, du \right)}_\text{visibility of $\ray(t)$ from $\origin$} \:\: \cdot \!\!\!\! \underbrace{\density(t)  \vphantom{\int_{t_n}^{t}} }_\text{density at $\ray(t)$}
\label{eq:weights}
\end{align}
where the volumetric density $\density(t)$ and radiance fields $\radiance(t)$ are predicted by a multi layer perceptron (i.e.~MLP) with Fourier feature encoding.
We refer the reader to the original work~\cite{Mildenhall20eccv_nerf} for further details and the discretization of these integrals~\cite{Max1995tvcg_VolumeRendering}.

\paragraph{Training}
While neural radiance fields are acknowledged to be slow to train, we find that we are able to fit a single model to sufficient quality in ${\approx}20$~minutes on eight TPUv3 cores on the Google Cloud Platform.
Once trained, the per scene parameters~$\{ \theta_\iScene \}$ are held fixed.

\subsection{Semantic Reasoning}
\label{sec:semantic}
We now present a method for mapping 3D density fields to 3D semantic fields.
To recap, we train a \textit{translation} model $\translator(\density \given \NesfParams)$ to produce a semantic field $\semfield(\point \given \density, \NesfParams)$ being given access to the density field of a scene $\density$, where $\semfield$ assigns a probability distribution over semantic categories at each 3D point in space.
We optimize translation model's parameters $\NesfParams$ with 2D annotations alone.
Our inspiration is drawn from methods that translate \textit{explicit} representations of geometry into semantics\cite{qi2017pointnet,dai2018scancomplete}, and in observing that for density fields provide an \textit{implicit} notion of geometry.
For ease of exposition, the overall architecture illustrate in \Figure{overview}, is broken down into several discrete steps:
\CIRCLE{1} density grid extraction,
\CIRCLE{2} spatial reasoning,
\CIRCLE{3} feature decoding,
\CIRCLE{4} supervision, and
\CIRCLE{5} data augmentation.

\paragraph{Density grid extraction}
Our method starts by uniformly evaluating the density field on a 3D lattice with spacing $\latticeResolution$ between samples:
\begin{equation}
\densityGrid_s 
= \densityGrid(\density_{\iScene}) 
= \{ \density(\point \given \theta_\iScene) \text{~s.t.~} \point \in [-1:\latticeResolution:+1]^3 \}
\end{equation}
While this operation limits the spatial resolution of the original density field, it presents a natural representation for further processing.

\paragraph{Spatial reasoning (3D)}
We apply a 3D~UNet~\cite{cciccek20163d} to $\densityGrid_\iScene$ to obtain a \textit{feature} grid $\featureGrid_s$ of the same spatial resolution as~$\densityGrid_\iScene$:
\begin{align}
\featureGrid_\iScene = \featureGrid(\densityGrid_{\iScene} | \NesfParams_{\unet})= \text{UNet3D}(\densityGrid_\iScene \given \NesfParams_{\unet})
\end{align}
This step is essential as a point-wise measurement of the density field $\density(\point)$ does not contain sufficient information to capture 3D structure.
After all, $\density(\point)$ only measures the volumetric density at a point, while the 3D structure requires reasoning over local spatial neighborhoods.
Note that we \emph{share} translation network parameters $\NesfParams_{\unet}$ across scenes, enabling generalization to novel scenes not available at training time.

\paragraph{Feature decoding}
\newcommand{\decoder}{\mathcal{D}}
Given a query point~$\point \in \RealNumbers^3$, we interpolate within the feature grid $\featureGrid_{\iScene}$ to obtain a feature vector corresponding to~$\point$.
We then employ a neural network decoder~$\decoder$ to generate a field of probability distributions over semantic categories:
\begin{equation}
\semfield(\point \given \featureGrid_\iScene, \NesfParams_{\mlp}) = \decoder( \TriLerp(\point, \featureGrid_{\iScene}) \given \NesfParams_{\mlp})
\end{equation}
where $\decoder$ is a multilayer perceptron with trainable parameters $\NesfParams_{\mlp}$ and \TriLerp{} applies trilinear interpolation similar to~\cite{Liu20neurips_sparse_nerf}.
Like their UNet counterpart, parameters $\NesfParams_{\mlp}$ are \textit{shared} across all scenes.

\paragraph{Supervision}
To supervise the training of parameters $\NesfParams$,  we employ volumetric rendering as in \NeRF{}~\cite{Mildenhall20eccv_nerf} but adapt it to render semantic maps as in~Semantic-NeRF~\cite{Zhi21iccv_SemanticNeRF}: 
\begin{align}
\semantic(\ray \given \density_\iScene, \NesfParams) = \int_\near^\far 
w(t \given \density_\iScene)
\cdot
\semfield(\point \given \density_\iScene, \NesfParams)  \, dt
\label{eq:semantic}
\end{align}
We supervise the training process for $\NesfParams$ by minimizing the softmax cross-entropy between rendered semantic and ground truth semantic maps along with a smoothness regularization term:
\begin{align}
\loss{sem}(\NesfParams) &= \expect_\iScene \left[ \loss{sem}(\theta_\iScene, \NesfParams) \right] \quad \text{where:}
\\
\loss{sem}(\theta_\iScene, \NesfParams) &= \sum_\iCamera \expect_{\ray \sim \Rays(\camera_\iCamera)}
\left[ \text{CE}(\semantic(\ray \given \theta_\iScene, \NesfParams) - \semantic^\gt_\iCamera(\ray)) \right]
\nonumber
\end{align}
We include an additional smoothness regularization term to encourage similar predictions in local neighborhoods. We sample points $\point \sim \text{Uniform}([-1, 1]^3)$ and normally distributed noise $\epsilon \sim \mathcal{N}(0, 0.01)$,
\begin{align}
\loss{reg}(\NesfParams) &= \expect_{\point, \epsilon} \left[ 
    || \semfield(x|\featureGrid, \NesfParams) - 
       \semfield(x + \epsilon|\featureGrid, \NesfParams) ||_2^2
\right]
\label{eq:loss_regularization}
\end{align}
Our total loss is thus $\loss{}(\NesfParams) = \loss{sem}(\NesfParams) + \lambda \loss{reg}(\NesfParams)$.

\paragraph{Data augmentation}
To increase the robustness of our method and similarly to classical methods~\cite{qi2017pointnet, qi2017pointnet++}, we apply data augmentation in the form of random rotations around the z-axis (i.e.~upwards).
In particular, we randomly sample an angle $\Angle \in [0, 2\pi)$ at each step of training.
Rather than extracting a density grid in \NeRF{}'s original coordinate system, we construct a rotation transformation $\RandRot_{\Angle}$ and query \NeRF{} at points $\point' = \RandRot_{\Angle}(\point)$, resulting in the following density grid:
\begin{align}
\tilde{\densityGrid}_{\iScene}
&= \{ \density(\point' \given \theta_\iScene) \text{~s.t.~} \point \in \RandRot^{-1}_{\Angle}([-1:\latticeResolution:+1]^3)) \}
\end{align}
Note that this procedure does not necessitate the retraining of \NeRF{} models.
\begin{table}
\footnotesize
\centering
\resizebox{\linewidth}{!}{ %
\begin{tabular}{l|c|c|c} 
\toprule
  & \KLEVR
  & \ToyBoxEasy
  & \ToyBoxHard
  \\
\midrule
\# scenes
  & 100 / 20
  & 500 / 25
  & 500 / 25
  \\
\shortstack{\# cameras/scene}
  & 210 / 90
  & 210 / 90
  & 210 / 90
  \\
\# total cameras
  & 36,000    %
  & 1,575,00  %
  & 1,575,00  %
  \\
\shortstack{frame resolution}
  & 256 $\times$ 256
  & 256 $\times$ 256
  & 256 $\times$ 256
  \\
\shortstack{\# objects/scene}
  & 4-12
  & 4-12
  & 4-12
  \\
\# object instances
  & 5
  & 25,905
  & 39,695
  \\
\# background instances
  & 1
  & 383
  & 383
  \\
\bottomrule
\end{tabular}
} %
\vspace{-1em}
\caption{
\textbf{Dataset statistics} -- 
Each dataset consists of a set of train and novel scenes, wherein each scene's cameras are split into a train and test set (denoted by a ``/'').
}
\label{tab:datasets}
\end{table}

\section{Datasets -- \Table{datasets} and \Figure{dataset_examples}}
\label{sec:datasets}
To investigate \NeSF{}, we require datasets describing the appearance and semantics of a large number of scenes from multiple points of view.
While existing datasets based on indoor and self-driving sensor captures exist~\cite{dai2017scannet,chang2017matterport3d,caesar2020nuscenes,sun2020scalability}, we desire a controlled setting where distractors such as motion blur, camera calibration error, and object motion can be eliminated.~$\quad$~To this end, we introduce three new datasets built on Kubric~\cite{greff2021kubric}: \KLEVR{}, \ToyBoxEasy, and \ToyBoxHard.
Each dataset consists of hundreds of synthetic scenes, each containing randomly-placed 3D objects which are photo-realistically rendered by a path tracer~\cite{blender} supporting soft shadows, physically based materials, and global illumination effects.
Each scene is described by a set of posed frames, where each frame provides an RGB image, a semantic map, and a depth map rendered from a shared camera pose.
We provide the Kubric worker script to generate such scenes, so as to enable follow-up research to build ever-more-challenging datasets.

\paragraph{\KLEVR}
We design the \KLEVR{} dataset to be a simple testbed akin to MNIST in machine learning.
Inspired by CLEVR~\cite{johnson2017clevr}, each scene contains 4 to 12 simple geometric objects placed randomly on a matte grey floor.
Each object is assigned a random hue and scale, and is constrained to lie within a fixed bounding box.
The semantic category of an object is set equal to its geometry class (e.g. cube, cylinder, etc).
While only the shape of each object is semantically relevant, color, scale, and placement serve as distractors.
For each scene, we render 300 frames from randomly-sampled camera poses aimed at the scene's origin. 
Camera poses are constrained to lie in the upper hemisphere surrounding the scene.
For each frame, we render an RGB image, a semantic map, and a depth map.

\newcommand{\minifigure}[2]{\includegraphics[width=6em]{#2}}
\begin{figure}
\small
\resizebox{\linewidth}{!}{ %< auto-adjusts font size to fill line
\begin{tabular}{ccc}
\KLEVRShort{}
    & \ToyBoxEasy
    & \ToyBoxHard
    \\
\minifigure{\KLEVR}{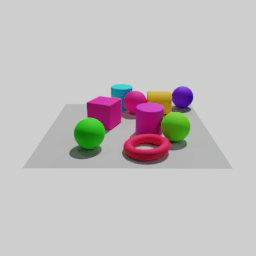}
  & \minifigure{\ToyBoxEasy}{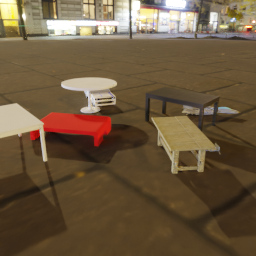}
  & \minifigure{\ToyBoxHard}{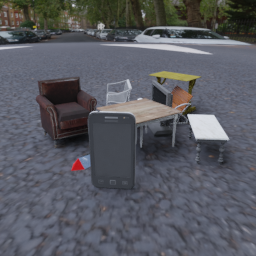}
  \\
\minifigure{}{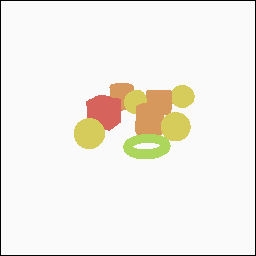}
  & \minifigure{}{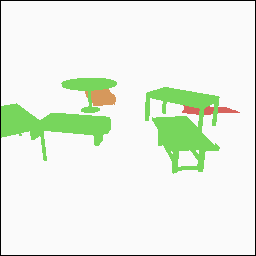}
  & \minifigure{}{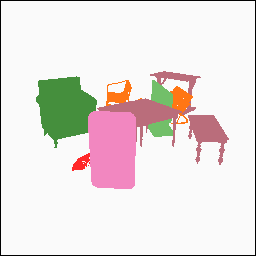}
  \\
\end{tabular}
} %< \resizebox
\vspace{-1em}
\caption{
\textbf{Dataset examples} -- Each frame includes an RGB image, semantic map, and depth map~(not pictured here).
}
\label{fig:dataset_examples}
\end{figure}

\paragraph{\ToyBoxEasy and \ToyBoxHard}
These datasets are designed to imitate scenes of children's bedrooms and are designed to be more challenging.
Scenes are constructed from a large vocabulary of ShapeNet~\cite{chang2015shapenet} objects coupled with HDRI backdrops~(floor, horizon, and environment illumination) captured in the wild~\cite{hdrihaven}.
Like \KLEVR{}, each scene consists of 4-12 randomly-placed objects, and frames are rendered from 300 independently-sampled camera poses.  
Objects are sampled at random from the $5$ and $13$ most common object categories, respectively for \ToyBoxEasy and \ToyBoxHard.
Such splits have been commonly used in the 3D deep learning literature~\cite{Groueix18cvpr_AtlasNet, Mescheder19cvpr_occupancy_net, Genova19iccv_sif, Deng20cvpr_cvxnet}.
Like the objects themselves, backdrops are sampled at random when constructing a scene.
With thousands of objects per category to choose from, most object instances appear rarely or only once.
 
\paragraph{Train/Test splits}
To enable evaluation from novel views within the same scene, we randomly partition each scene's frames into \TrainViews and \TestViews; the latter representing the set typically used to evaluate methods in novel view synthesis~\cite{Mildenhall20eccv_nerf}.
For evaluation \emph{across} scenes, we further partition scenes into \TrainScenes  and \TestScenes.

\section{Experiments}
\label{sec:experiments}

We evaluate \NeSF{} on the three datasets described in Section~\ref{sec:datasets}.
Unless otherwise stated, we train \NeRF{} models on all \TrainViews from all \TrainScenes.
To mimic a label-scarce regime, we choose to provide \NeSF{} supervision from semantic maps corresponding to \NumTrainImagesPerScene randomly-chosen cameras per scene.
For 2D evaluation, we randomly select \NumTestImagesPerScene cameras from each \TestScenes' \TrainViews.
For 3D evaluation, we use camera parameters and ground truth depth maps to derive a labeled 3D point cloud from the same \NumTestImagesPerScene cameras.
Semantic segmentations are evaluated according to 2D and 3D mean intersection-over-union.
Further details are provided in the \SupplementaryMaterial.

\paragraph{Training Details}
Each scene is preprocessed by training an independent \NeRF{} for $25k$ steps with Adam using an initial learning rate of $1e{-}3$ decaying to $5.4e{-}4$ according to a cosine rule.
Our \NeRF{} architecture follows the original work.
\NeSF{} is trained for $5k$ steps using Adam with an initial learning rate of $1e{-}3$ decaying to $4e{-}4$.
As input for \NeSF{}, we discretize density fields by densely probing with $\epsilon{=}1/32$ resulting in $64^3$ evenly-spaced points in $[-1,+1]^3$.
This density grid is then processed by the 3D UNet architecture of {\c{C}}i{\c{c}}ek~ \textit{et al.}~\cite{cciccek20163d} with 32, 64, and 128 channels at each stage of downsampling.
The semantic latent vector is processed by a multilayer perceptron consisting of 2 hidden layers of 128 units. Our models are trained on 32 TPUv3 cores.

\paragraph{Segmentation baselines (2D/3D)}
We compare \NeSF{} to two popular semantic segmentation baselines, \DeepLab{}~\cite{chen2017deeplab} and \SparseConvNet{}~\cite{graham2017submanifold}.
\DeepLab{} follows a traditional 2D semantic segmentation pipeline, producing segmentation maps from RGB images.
We train \DeepLab{} v3 with Wide ResNet~\cite{wu2019wider} for $55k$ steps on 16 TPUv3 chips.
\SparseConvNet{} is a point cloud segmentation method and, unlike \NeSF{} and \DeepLab{}, requires explicit 3D supervision.
We train \SparseConvNet{} asynchronously on 20 NVIDIA V100 GPUs with momentum using a base learning rate of $1.5e{-}2$ and decaying to $0$ over the final $250k$ steps of training.
We refer the reader to \Section{comparison-to-baselines} and the \SupplementaryMaterial{} for further details.

\begin{table}
\centering
\resizebox{\linewidth}{!}{ %
\begin{tabular}{lcc|cc} 
\toprule
& \multicolumn{2}{c|}{\TrainViews{}} 
    & \multicolumn{2}{c}{\TestViews{}}  
    \\
& 2D mIoU          & 3D   mIoU             
    & 2D mIoU          & 3D mIoU   
    \\
\hline
\NeSF{}                           
    & $92.7$  & $97.8$  
    & $92.6$  & $97.5$ 
    \\
\DeepLab{}\cite{chen2017deeplab}     
    & $97.1$          & N/A                    
    & N/A               & N/A  
    \\
\rowcolor[gray]{0.9}
\SparseConvNet{}~\cite{graham2017submanifold}
    & N/A               & $99.7$              
    & N/A               & $99.7$ 
    \\
\bottomrule
\end{tabular}
}
\caption{
    \textbf{Quantitative comparison on \KLEVR{}} -- 
    \NeSF{} is competitive with 2D and 3D baselines.
    See additional details in~\Table{sota_multi_shapenet}.
}
\label{tab:sota_klevr}
\end{table}

\begin{table*}
\centering
\resizebox{\linewidth}{!}{ %
\begin{tabular}{lcc|cc|cc|cc} 
\toprule
& \multicolumn{4}{c|}{\ToyBoxEasy}
    & \multicolumn{4}{c}{\ToyBoxHard}   \\
\cline{2-9}
& \multicolumn{2}{c|}{\TrainViews{}} & \multicolumn{2}{c|}{\TestViews{}}
    & \multicolumn{2}{c|}{\TrainViews{}} & \multicolumn{2}{c}{\TestViews{}}   \\
& 2D \mIoU{}         & 3D \mIoU{}              & 2D \mIoU{}         & 3D \mIoU{}
    & 2D \mIoU{}         & 3D \mIoU{}              & 2D \mIoU{}         & 3D \mIoU{}        \\
\hline
\NeSF{}
& $81.9 \pm 0.8$ & $88.7 \pm 0.9$ & $81.7 \pm 0.6$  & $89.6 \pm 0.7$
    & $56.5 \pm 0.8$  & $60.1 \pm 0.6$  & $56.6 \pm 1.0$  & $61.9 \pm 0.9$ \\
\DeepLab{}\cite{chen2017deeplab}
& $81.6$          & N/A                    & N/A               & N/A
    & $63.1$          & N/A                    & N/A               & N/A              \\
\rowcolor[gray]{0.9}
\SparseConvNet{}\cite{graham2017submanifold}
& N/A               &  $93.4$              & N/A               & $94.8$
    & N/A               &  $83.2$              & N/A               & $81.7$ \\
\bottomrule
\end{tabular}
} %
\vspace{-.5em}
\caption{
    \textbf{Quantitative comparison} -- 
    \NeSF{} is competitive with 2D and 3D baselines.
    At train time, \NeSF{} and \DeepLab{} only utilize 2D supervision.
    Conversely, \SparseConvNet{} requires full 3D supervision in the form of labeled 3D point clouds.
    We construct oracle point clouds via back-projected depth maps, resulting in an upper bound to our method (grayed-out row).
    Models are evaluated on train and test camera poses from test scenes.
    Configurations marked as ``N/A'' denotes a setting where methods are not applicable.
    Statistics for \NeSF{} are aggregated across five random initializations.
}
\label{tab:sota_multi_shapenet}
\end{table*}

\subsection{Comparisons to baselines}
\label{sec:comparison-to-baselines}
Our first set of experiments evaluates the performance of our proposed method in comparison to alternative benchmark methods on the \KLEVR{}, \MultiShapeNet{}-5, and \MultiShapeNet{}-13 datasets.
As far as we are aware, \NeSF{} is the \textit{first} method capable of simultaneously producing 3D geometry, 2D semantic maps, and 3D semantic labels directly from posed RGB images at inference time.
Unlike prior work~\cite{Murez20eccv_atlas}, our method is trained on posed 2D supervision alone.
As no existing method is directly comparable, we compare \NeSF{} to competitive baselines for 2D image segmentation and 3D point cloud segmentation.

\paragraph{Comparison to \DeepLab{}~\cite{chen2017deeplab} (2D)}
To maintain a fair comparison in 2D, we train both the semantic phase of \NeSF{} and \DeepLab{} on an identical set of paired RGB images and semantic maps for a fixed set of scenes (i.e.~\NumTrainImagesPerScene per scene in \TrainScenes).
\NeSF{} has further access to all \NumRgbImagesPerScene RGB maps associated with each scene's \TrainViews, which are used to fit per-scene \NeRF{} models.
Both methods are evaluated on a random sampling of frames from \TestScenes, \NumTestImagesPerScene per scene.
To emphasize the 3D nature of \NeSF{}, we evaluate on additional camera poses from \TestScenes where RGB information is not available, an additional \NumTestImagesPerScene per scene from each scene's \TestViews.

\paragraph{Comparison to \SparseConvNet{}~\cite{graham2017submanifold} (3D)}
As \SparseConvNet{} requires 3D input, we derive an oracle point cloud for each scene from camera poses and \textit{ground truth} depth maps -- hence giving an \textit{unfair advantage} to this method, establishing an upper bound on performance given full 3D supervision.
For this, we use the same \NumRgbImagesPerScene train frames used to fit \NeRF{} models in the 2D comparison.
We further select a subset of each point cloud for 3D semantic supervision; namely, the points corresponding to the \NumTrainImagesPerScene semantic maps supervising \NeSF{} and \DeepLab{}.
We evaluate \NeSF{} and \SparseConvNet{} on two sets of 3D points on \TestScenes.
The first set is a subset of each point cloud corresponding to \NumTestImagesPerScene randomly chosen frames from each scene's \TrainViews.
These points are available to \SparseConvNet{} as part of each scene's 3D representation.
The second set is a set of additional query points derived from \NumTestImagesPerScene additional frames from each scene's \TestViews.
As \SparseConvNet{} is not designed to classify points beyond its input point cloud, we apply a nearest neighbor label propagation procedure to assign labels to the latter.

\paragraph{Quantitative comparisons -- \Table{sota_klevr} and \Table{sota_multi_shapenet}}
While all methods perform comparably on the \KLEVR{} dataset, model quality varies drastically on the more challenging datasets.
On \ToyBoxEasy, our method performs comparably to \DeepLab{}, but on \ToyBoxHard{}, it underperforms by 6.6\% in 2D mIoU.
While our method does not achieve the same level of accuracy as \DeepLab{} on frames where RGB images are available, it is able to achieve near identical accuracy on \emph{novel} camera poses, a task \DeepLab{} is unable to approach.
In order to focus on the foundational properties of our method, we have chosen to limit \NeSF{} to 3D geometric information alone.
Incorporation of 2D information via projection onto ground truth cameras in the spirit of PixelNeRF~\cite{Yu21cvpr_pixelNeRF} or IBRNet~\cite{Wang21cvpr_IBRNet} is straightforward.

As expected, our method also underperforms \SparseConvNet{} by 4.7-5.2\% on \ToyBoxEasy and 19.8-23.1\% on \ToyBoxHard.
Unlike \SparseConvNet{}, our method lacks access to dense, ground truth depth maps and full 3D supervision.
Further, the 3D UNet architecture employed by \NeSF{} is based on~\cite{cciccek20163d}, a predecessor to the \SparseConvNet{} architecture.
As \NeSF{} does not take advantage of sparsity, it must operate at a lower spatial resolution than the baseline and tends to mislabel small objects and thin structures.
Though \NeSF{} underperforms \SparseConvNet{} today, we anticipate methodological improvements in model architecture to rapidly improve performance.
Additional in-depth analysis is included in the \SupplementaryMaterial. 

\begin{figure*}

\setlength\tabcolsep{0pt}
\def\arraystretch{1.0}

{  % 
\newcommand{\includeimage}[1] {
  \includegraphics[width=3cm]{fig/qualitative/#1}
}

\newcommand{\includeimageset}[1]{
\includeimage{#1/rgb}
  & \includeimage{#1/label}
  & \includeimage{#1/nesf}
  & \includeimage{#1/deeplab}
  & \includeimage{#1/sparseconvnet}
}

\newcommand{\insertlabel}[1]{
    #1
}
\centering

% -- Main table
% \resizebox{\linewidth}{!}{ %< auto-adjusts font size to fill line
    \begin{tabular}{ccccc}
        \includeimageset{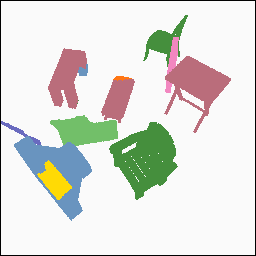} \\
        \includeimageset{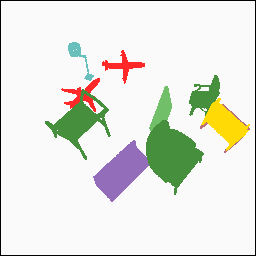} \\
        \includeimageset{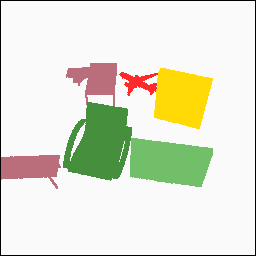} \\
        \insertlabel{RGB}
            & \insertlabel{Ground Truth}
            & \insertlabel{\NeSF{} (Ours)}
            & \insertlabel{\DeepLab{}~\cite{chen2017deeplab}}
            & \insertlabel{\SparseConvNet{}~\cite{graham2017submanifold}}
    \end{tabular}
% } %< \resizebox
% \vspace{-1em}
\caption{
\textbf{Qualitative comparison (\ToyBoxHard)} --
Unlike \DeepLab{}, \NeSF{} is able to clearly separate objects with similar appearance but different geometry (Top).
However, \NeSF{} struggles with thin structures like lamp posts (Middle) and tends to smear labels from nearby objects (Bottom).
\SparseConvNet{} suffers from neither limitation but has access to oracle 3D geometry and full 3D supervision.
}
\label{fig:qualitative_multi_shapenet}
}
\end{figure*}

\paragraph{Qualitative comparisons -- \Figure{qualitative_multi_shapenet}}
Qualitatively, our method exhibits strong performance in identifying the 13 canonical categories in \ToyBoxHard{}.
Because our method operates directly on 3D geometry, it is not easily confused by objects of similar appearance but dissimilar geometry as demonstrated by the thin standing rifle in the top row.
While \NeSF{} and \SparseConvNet{} correctly recognize the rifle's geometry, \DeepLab{} labels it identically to the chair behind it.~$\quad$~
Similar to \DeepLab{}, our method faces challenges with thin structures such as the tube of the standing lamp pictured in the middle row. 
Such structures are not well captured by the regular grids employed by \DeepLab{} and \NeSF{} in 2D and 3D, respectively.
With access to dense, accurate point clouds, \SparseConvNet{} correctly identifies the lamp in its entirety.~$\quad$~One limitation particularly evident in \NeSF{} is a tendency to smear labels across nearby objects, as demonstrated by the chair partially labeled as a display in the bottom row.
Without access to \textit{appearance} or fine-grained geometry, \NeSF{} is unable to identify when one object ends and another begins.
Integrating appearance information and access to higher spatial resolution are straightforward methods for improving \NeSF{}'s accuracy in future work.

\begin{table}
\small
\centering
\begin{tabular}{c|c|c|c} 
\toprule
\multicolumn{2}{c|}{Hyperparameter}
  & 2D & 3D \\
\hline
\multirow{2}{*}{\shortstack{Random \\ Rotations}}
  & No
  & 81.1  & 75.5 \\
  & Yes
  & 92.0  & 97.1 \\
\hline

\multirow{4}{*}{\shortstack{Density \\ Grid}}
  & (32, 32, 32)
  & 87.5  & 92.1 \\
  & (48, 48, 48)
  & 91.2  & 96.0 \\
  & (64, 64, 64)
  & 91.7  & 89.6 \\
  & (80, 80, 80)
  & 92.0  & 97.1 \\
\hline

\multirow{3}{*}{UNet}
  & (16, 32, 64)
  & 89.9  & 94.4 \\
  & (24, 48, 96)
  & 91.5  & 96.4 \\
  & (32, 64, 128)
  & 92.0  & 97.1 \\
\hline

\multirow{4}{*}{MLP}
  & (0, 32)
  & 91.3  & 96.4 \\
  & (1, 32)
  & 91.8  & 96.9 \\
  & (1, 64)
  & 91.2  & 96.2 \\
  & (2, 128)
  & 92.0  & 97.1 \\
  
\bottomrule
\end{tabular}
\caption{
    \textbf{Ablation: hyper-parameters} --
    Data augmentation, in the form of random scene rotations, increased spatial resolution of the density grid, and increased UNet model capacity improve 2D and 3D \mIoU{}.
    Experiments on 25 scenes from the \KLEVR{} dataset.
}
\label{tab:nesf_ablations}
\end{table}

\begin{table}
\small
\centering
\begin{tabular}{ccc|cc}
\toprule
\multirow{2}{*}{\shortstack{\# RGB \\ Images}}
  & \multicolumn{2}{c|}{\NeRF{}}
  & \multicolumn{2}{c}{\NeSF{}} \\
& PSNR & SSIM
  & 2D & 3D \\
\hline
5    & $21.2 \pm 1.4$   & $0.89 \pm 0.02$   & $52.2$   & $72.4$  \\
10   & $24.2 \pm 1.2$   & $0.92 \pm 0.01$   & $79.6$   & $96.7$  \\
25   & $30.3 \pm 1.2$   & $0.96 \pm 0.01$   & $87.3$   & $97.0$  \\
50   & $35.5 \pm 0.9$   & $0.98 \pm 0.00$   & $90.8$   & $97.3$  \\
75   & $37.5 \pm 1.0$   & $0.98 \pm 0.00$   & $91.4$   & $97.1$  \\
100  & $38.4 \pm 1.1$   & $0.98 \pm 0.00$   & $92.0$   & $97.4$  \\
\bottomrule
\end{tabular}
\caption{
\textbf{Ablation: sensitivity to reconstruction quality} --
The accuracy of our method improves with \NeRF{}'s reconstruction quality.
PSNR and SSIM are are averaged across all scenes and metrics aggregated. Experiments on all scenes from \KLEVR{}.
}
\label{tab:reconstruction_quality}
\end{table}

% Figure generated by Colab,
% https://colab.corp.google.com/drive/1vbeqGF63xHScQzzeHi4c5LUGWURV8bQ2?resourcekey=0-j_6KEX0t08UWOryMF-IHag#scrollTo=5xkUL1E7ijs2&line=26&uniqifier=1

\begin{figure}[t]
\centering
\includegraphics[width=0.9\linewidth]{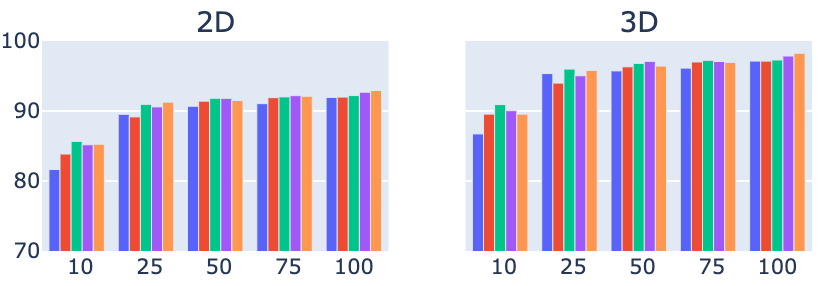}
\vspace{-.5em}
\caption{
\textbf{Ablation: data efficiency} --
2D and 3D mIoU as a function of the number of train scenes for scenes with supervision from~{\color{blue} \textbf{1}},~{\color{red} \textbf{2}},~{\color{ForestGreen} \textbf{5}},~{\color{purple} \textbf{10}}, or~{\color{YellowOrange} \textbf{25}}~semantic maps per scene.
\NeSF{} generalizes to new scenes with as few as a \emph{one semantic map per scene}.
Additional semantic maps per scene marginally improve the accuracy.
Experiments on \KLEVR{} dataset.
}
\label{fig:num_images_per_scene}
\end{figure}

\subsection{Ablation Studies}
Our second set of experiments investigates how each component of our method affects system performance on the \KLEVR{} dataset.

\paragraph{Sensitivity to features}
Table~\ref{tab:nesf_ablations} shows results of an ablation study where one feature of our method is varied while holding all others to their reference values.
We find that each component provides a measurable improvement in 2D and 3D segmentation quality.
Data augmentation in the form of random scene rotations improves quality the most, adding 10.3\% and 11.8\% to 2D and 3D \mIoU{} respectively.
The spatial resolution of the probed \NeRF{} density grids is the second most important as insufficient resolution makes smaller objects indistinguishable.

\paragraph{Sensitivity to reconstruction quality -- \Table{reconstruction_quality}}
We investigate the robustness of \NeSF{} to \NeRF{} reconstruction quality.
To modulate reconstruction quality, we vary the number of RGB images used when fitting \NeRF{} models from 5 to 100.
As expected, \NeRF{} reconstruction quality as measured on novel views increases as more RGB images are provided.
At the same time, we find that \NeSF{} 2D segmentation accuracy improves monotonically with \NeRF{} reconstruction quality.
Improvement of NeRF reconstruction quality is a highly active area of research, as such we anticipate methodological improvements may be directly applied to improve NeSF's performance. 
Surprisingly, 3D segmentation accuracy levels off near 97\% when \NeRF{} models are optimized with as little as 25 RGB images.

\boldpnov{Sensitivity to data scarcity -- \Figure{num_images_per_scene}}
To investigate \NeSF{}'s applicability to scenarios where labeled semantic maps are scarce, we investigate robustness to the number of semantic maps per scene.
We find \NeSF{} easily generalizes to novel scenes with \emph{as few as one semantic map per train scene}.
Additional semantic maps per scene improve performance given a small number of scenes, with no noticeable effect after 25 scenes.
This suggests datasets consisting of videos, each with a single labeled frame, are ideal for \NeSF{}.
\section{Conclusions and Limitations}

In this work, we present \NeSF{}, a novel method for simultaneous 3D scene reconstruction and semantic segmentation from posed 2D images.
Building on \NeRF{}, our method is trained solely on posed 2D RGB images and semantic maps.
At inference time, our method constructs a dense semantic segmentation field that can be queried directly in 3D or used to render 2D semantic maps from novel camera poses.
We compare \NeSF{} to competitive baselines in 2D and 3D semantic segmentation on three novel datasets.

In more challenging settings, we find that \NeSF{} underperforms its baselines.
However, \NeSF{} offers novel capabilities.
Unlike traditional 2D segmentation methods, \NeSF{} fuses information across multiple independent views and renders semantic maps from novel poses.
Unlike 3D point cloud methods, \NeSF{} operates on posed 2D information alone at both train and test time.
We chose to limit \NeSF{} to a core set of features to better explore the fundamental trade-offs and capabilities of such an approach. We further explore trade offs with respect to model choice, and potential social impact in the \SupplementaryMaterial.
In future work, we anticipate extending \NeSF{} to incorporate 2D semantic models and 3D sparsity will significantly improve accuracy. 

In addition to \NeSF{}, we propose three new datasets for multiview 3D reconstruction and semantic segmentation totalling over 3,000,000 frames and 1,000 scenes.
Each dataset contains hundreds of scenes, each consisting of a set of randomly placed objects.
The more challenging of these datasets are rendered with realistic illumination and a large catalogue of objects and backgrounds.
These datasets along with accompanying code and pretrained \NeRF{} models, will be released to the public upon publication.

\section*{Acknowledgements}

\dd{
We would like to express our gratitude to Konstantinos Rematas, D. Sculley, and especially Thomas Funkhouser for their ideas and suggestions. 
We would like to note our deep appreciation for the project support and leadership given by Jakob Uszkoreit, without which the work would not have been possible. 
The authors would also like to thank Rocky Cai in his assistance in assembling the \DeepLab{} baseline experiments.
}

{
    \small
    \bibliographystyle{ieee_fullname}
    \bibliography{macros,nerf,3dseg,main}

\begin{thebibliography}{100}\itemsep=-1pt

\bibitem{alwassel2020self}
Humam Alwassel, Dhruv Mahajan, Bruno Korbar, Lorenzo Torresani, Bernard Ghanem,
  and Du Tran.
\newblock Self-supervised learning by cross-modal audio-video clustering.
\newblock {\em Advances in Neural Information Processing Systems}, 33, 2020.

\bibitem{armeni_iccv19}
Iro Armeni, Zhi-Yang He, JunYoung Gwak, Amir~R. Zamir, Martin Fischer, Jitendra
  Malik, and Silvio Savarese.
\newblock 3d scene graph: A structure for unified semantics, 3d space, and
  camera.
\newblock In {\em Proceedings of the IEEE International Conference on Computer
  Vision}, 2019.

\bibitem{aytar2016soundnet}
Yusuf Aytar, Carl Vondrick, and Antonio Torralba.
\newblock Soundnet: Learning sound representations from unlabeled video.
\newblock In {\em Advances in neural information processing systems}, pages
  892--900, 2016.

\bibitem{badrinarayanan2017segnet}
Vijay Badrinarayanan, Alex Kendall, and Roberto Cipolla.
\newblock Segnet: A deep convolutional encoder-decoder architecture for image
  segmentation.
\newblock {\em IEEE transactions on pattern analysis and machine intelligence},
  39(12):2481--2495, 2017.

\bibitem{Barron21iccv_Mip_NeRF}
Jonathan Barron, Ben Mildenhall, Matthew Tancik, Peter Hedman, Ricardo
  Martin-Brualla, and Pratul Srinivasan.
\newblock {Mip-NeRF: A Multiscale Representation for Anti-Aliasing Neural
  Radiance Fields}.
\newblock In {\em ICCV}, October 2021.

\bibitem{billard2019trends}
Aude Billard and Danica Kragic.
\newblock Trends and challenges in robot manipulation.
\newblock {\em Science}, 364(6446), 2019.

\bibitem{bonin2008visual}
Francisco Bonin-Font, Alberto Ortiz, and Gabriel Oliver.
\newblock Visual navigation for mobile robots: A survey.
\newblock {\em Journal of intelligent and robotic systems}, 53(3):263--296,
  2008.

\bibitem{Boss21iccv_NeRD}
Mark Boss, Raphael Braun, Varun Jampani, Jonathan Barron, Ce Liu, and Hendrik
  Lensch.
\newblock {NeRD: Neural Reflectance Decomposition from Image Collections}.
\newblock In {\em ICCV}, October 2021.

\bibitem{caesar2019nuscenes}
Holger Caesar, Varun Bankiti, Alex~H Lang, Sourabh Vora, Venice~Erin Liong,
  Qiang Xu, Anush Krishnan, Yu Pan, Giancarlo Baldan, and Oscar Beijbom.
\newblock {nuScenes}: A multimodal dataset for autonomous driving.
\newblock In {\em Proceedings of the IEEE/CVF conference on computer vision and
  pattern recognition}, pages 11621--11631, 2020.

\bibitem{caesar2020nuscenes}
Holger Caesar, Varun Bankiti, Alex~H Lang, Sourabh Vora, Venice~Erin Liong,
  Qiang Xu, Anush Krishnan, Yu Pan, Giancarlo Baldan, and Oscar Beijbom.
\newblock nuscenes: A multimodal dataset for autonomous driving.
\newblock In {\em Proceedings of the IEEE/CVF conference on computer vision and
  pattern recognition}, pages 11621--11631, 2020.

\bibitem{Chabra20eccv_DLS}
Rohan Chabra, Jan Lenssen, Eddy Ilg, Tanner Schmidt, Julian Straub, Steven
  Lovegrove, and Richard Newcombe.
\newblock {Deep Local Shapes: Learning Local SDF Priors for Detailed 3D
  Reconstruction}.
\newblock In {\em ECCV}, 2020.

\bibitem{chang2017matterport3d}
Angel Chang, Angela Dai, Thomas Funkhouser, Maciej Halber, Matthias Niessner,
  Manolis Savva, Shuran Song, Andy Zeng, and Yinda Zhang.
\newblock Matterport3d: Learning from rgb-d data in indoor environments.
\newblock {\em arXiv preprint arXiv:1709.06158}, 2017.

\bibitem{chang2015shapenet}
Angel~X Chang, Thomas Funkhouser, Leonidas Guibas, Pat Hanrahan, Qixing Huang,
  Zimo Li, Silvio Savarese, Manolis Savva, Shuran Song, Hao Su, et~al.
\newblock Shapenet: An information-rich 3d model repository.
\newblock {\em arXiv preprint arXiv:1512.03012}, 2015.

\bibitem{che2019object}
Erzhuo Che, Jaehoon Jung, and Michael~J Olsen.
\newblock Object recognition, segmentation, and classification of mobile laser
  scanning point clouds: A state of the art review.
\newblock {\em Sensors}, 19(4):810, 2019.

\bibitem{Chen21iccv_MVSNeRF}
Anpei Chen and Zexiang Xu.
\newblock {MVSNeRF: Fast Generalizable Radiance Field Reconstruction From
  Multi-View Stereo}.
\newblock In {\em ICCV}, October 2021.

\bibitem{chen2017deeplab}
Liang-Chieh Chen, George Papandreou, Iasonas Kokkinos, Kevin Murphy, and Alan~L
  Yuille.
\newblock Deeplab: Semantic image segmentation with deep convolutional nets,
  atrous convolution, and fully connected crfs.
\newblock {\em IEEE transactions on pattern analysis and machine intelligence},
  40(4):834--848, 2017.

\bibitem{chen2017rethinking}
Liang-Chieh Chen, George Papandreou, Florian Schroff, and Hartwig Adam.
\newblock Rethinking atrous convolution for semantic image segmentation.
\newblock {\em arXiv preprint arXiv:1706.05587}, 2017.

\bibitem{Chen20cvpr_bsp}
Zhiqin Chen, Andrea Tagliasacchi, and Hao Zhang.
\newblock {BSP-Net: Generating Compact Meshes via Binary Space Partitioning}.
\newblock In {\em CVPR}, 2020.

\bibitem{chen2019imnet}
Zhiqin Chen and Hao Zhang.
\newblock Learning implicit fields for generative shape modeling.
\newblock In {\em Proceedings of the IEEE/CVF Conference on Computer Vision and
  Pattern Recognition}, pages 5939--5948, 2019.

\bibitem{Chen19cvpr_IM_NET}
Zhiqin Chen and Hao Zhang.
\newblock {Learning Implicit Fields for Generative Shape Modeling}.
\newblock In {\em CVPR}, 2019.

\bibitem{cheng20212}
Ran Cheng, Ryan Razani, Ehsan Taghavi, Enxu Li, and Bingbing Liu.
\newblock (af)2-s3net: Attentive feature fusion with adaptive feature selection
  for sparse semantic segmentation network.
\newblock In {\em Proceedings of the IEEE/CVF Conference on Computer Vision and
  Pattern Recognition}, pages 12547--12556, 2021.

\bibitem{chollet2017xception}
Fran{\c{c}}ois Chollet.
\newblock Xception: Deep learning with depthwise separable convolutions.
\newblock In {\em Proceedings of the IEEE conference on computer vision and
  pattern recognition}, pages 1251--1258, 2017.

\bibitem{choy2019minkowskinet}
Christopher Choy, JunYoung Gwak, and Silvio Savarese.
\newblock 4d spatio-temporal convnets: Minkowski convolutional neural networks.
\newblock In {\em Proceedings of the IEEE/CVF Conference on Computer Vision and
  Pattern Recognition}, pages 3075--3084, 2019.

\bibitem{choy20194d}
Christopher Choy, JunYoung Gwak, and Silvio Savarese.
\newblock 4d spatio-temporal convnets: Minkowski convolutional neural networks.
\newblock In {\em Proceedings of the IEEE Conference on Computer Vision and
  Pattern Recognition}, pages 3075--3084, 2019.

\bibitem{choy2019fully}
Christopher Choy, Jaesik Park, and Vladlen Koltun.
\newblock Fully convolutional geometric features.
\newblock In {\em Proceedings of the IEEE International Conference on Computer
  Vision}, pages 8958--8966, 2019.

\bibitem{cciccek20163d}
{\"O}zg{\"u}n {\c{C}}i{\c{c}}ek, Ahmed Abdulkadir, Soeren~S Lienkamp, Thomas
  Brox, and Olaf Ronneberger.
\newblock 3d u-net: learning dense volumetric segmentation from sparse
  annotation.
\newblock In {\em International conference on medical image computing and
  computer-assisted intervention}, pages 424--432. Springer, 2016.

\bibitem{Cole21iccv_RtS}
Forrester Cole, Kyle Genova, Avneesh Sud, Daniel Vlasic, and Zhoutong Zhang.
\newblock Differentiable surface rendering via non-differentiable sampling.
\newblock In {\em ICCV}, pages 6088--6097, 2021.

\bibitem{blender}
Blender~Online Community.
\newblock {\em Blender - a 3D modelling and rendering package}.
\newblock Blender Foundation, Stichting Blender Foundation, Amsterdam, 2018.

\bibitem{cortinhal2020salsanext}
Tiago Cortinhal, George Tzelepis, and Eren~Erdal Aksoy.
\newblock Salsanext: Fast, uncertainty-aware semantic segmentation of lidar
  point clouds for autonomous driving.
\newblock {\em arXiv preprint arXiv:2003.03653}, 2020.

\bibitem{dai2017scannet}
Angela Dai, Angel~X Chang, Manolis Savva, Maciej Halber, Thomas Funkhouser, and
  Matthias Nie{\ss}ner.
\newblock Scannet: Richly-annotated 3d reconstructions of indoor scenes.
\newblock In {\em Proceedings of the IEEE Conference on Computer Vision and
  Pattern Recognition}, pages 5828--5839, 2017.

\bibitem{dai20183dmv}
Angela Dai and Matthias Nie{\ss}ner.
\newblock 3dmv: Joint 3d-multi-view prediction for 3d semantic scene
  segmentation.
\newblock In {\em Proceedings of the European Conference on Computer Vision
  (ECCV)}, pages 452--468, 2018.

\bibitem{dai2018scancomplete}
Angela Dai, Daniel Ritchie, Martin Bokeloh, Scott Reed, J{\"u}rgen Sturm, and
  Matthias Nie{\ss}ner.
\newblock Scancomplete: Large-scale scene completion and semantic segmentation
  for 3d scans.
\newblock In {\em Proceedings of the IEEE Conference on Computer Vision and
  Pattern Recognition}, pages 4578--4587, 2018.

\bibitem{Deng20cvpr_cvxnet}
Boyang Deng, Kyle Genova, Sofien Bouaziz, Geoffrey Hinton, Andrea Tagliasacchi,
  and Soroosh Yazdani.
\newblock {CvxNet: Learnable Convex Decomposition}.
\newblock In {\em CVPR}, 2020.

\bibitem{Deng20eccv_NASA}
Boyang Deng, JP Lewis, Timothy Jeruzalski, Gerard Pons-Moll, Geoffrey Hinton,
  Mohammad Norouzi, and Andrea Tagliasacchi.
\newblock {NASA: Neural Articulated Shape Approximation}.
\newblock In {\em ECCV}, 2020.

\bibitem{DeVries21iccv_GSN}
Terrance DeVries.
\newblock {Unconstrained Scene Generation With Locally Conditioned Radiance
  Fields}.
\newblock In {\em ICCV}, October 2021.

\bibitem{douillard2011segmentation}
Bertrand Douillard, James Underwood, Noah Kuntz, Vsevolod Vlaskine, Alastair
  Quadros, Peter Morton, and Alon Frenkel.
\newblock On the segmentation of 3d lidar point clouds.
\newblock In {\em 2011 IEEE International Conference on Robotics and
  Automation}, pages 2798--2805. IEEE, 2011.

\bibitem{ephrat2018looking}
Ariel Ephrat, Inbar Mosseri, Oran Lang, Tali Dekel, Kevin Wilson, Avinatan
  Hassidim, William~T. Freeman, and Michael Rubinstein.
\newblock Looking to listen at the cocktail party: A speaker-independent
  audio-visual model for speech separation.
\newblock {\em ACM Trans. Graph.}, 37(4), July 2018.

\bibitem{gan2019self}
Chuang Gan, Hang Zhao, Peihao Chen, David Cox, and Antonio Torralba.
\newblock Self-supervised moving vehicle tracking with stereo sound.
\newblock In {\em Proceedings of the IEEE International Conference on Computer
  Vision}, pages 7053--7062, 2019.

\bibitem{Garbin21iccv_FastNeRF}
Stephan Garbin and Marek Kowalski.
\newblock {FastNeRF: High-Fidelity Neural Rendering at 200FPS}.
\newblock In {\em ICCV}, October 2021.

\bibitem{Genova19iccv_sif}
Kyle Genova, Forrester Cole, Daniel Vlasic, Aaron Sarna, William Freeman, and
  Thomas Funkhouser.
\newblock {Learning Shape Templates with Structured Implicit Functions}.
\newblock In {\em ICCV}, 2019.

\bibitem{genova2021learning}
Kyle Genova, Xiaoqi Yin, Abhijit Kundu, Caroline Pantofaru, Forrester Cole,
  Avneesh Sud, Brian Brewington, Brian Shucker, and Thomas Funkhouser.
\newblock Learning 3d semantic segmentation with only 2d image supervision.
\newblock {\em 3DV}, 2021.

\bibitem{girdhar2019distinit}
Rohit Girdhar, Du Tran, Lorenzo Torresani, and Deva Ramanan.
\newblock Distinit: Learning video representations without a single labeled
  video.
\newblock In {\em Proceedings of the IEEE International Conference on Computer
  Vision}, pages 852--861, 2019.

\bibitem{graham20183d}
Benjamin Graham, Martin Engelcke, and Laurens van~der Maaten.
\newblock 3d semantic segmentation with submanifold sparse convolutional
  networks.
\newblock In {\em Proceedings of the IEEE conference on computer vision and
  pattern recognition}, pages 9224--9232, 2018.

\bibitem{graham2017submanifold}
Benjamin Graham and Laurens van~der Maaten.
\newblock Submanifold sparse convolutional networks.
\newblock {\em arXiv preprint arXiv:1706.01307}, 2017.

\bibitem{greff2021kubric}
Klaus Greff and Andrea Tagliasacchi.
\newblock Kubric.
\newblock \url{https://github.com/google-research/kubric}, 2021.

\bibitem{Groueix18cvpr_AtlasNet}
Thibault Groueix, Matthew Fisher, Vladimir~G. Kim, Bryan~C. Russell, and
  Mathieu Aubry.
\newblock {AtlasNet: A Papier-Mâché Approach to Learning 3D Surface
  Generation}.
\newblock In {\em CVPR}, 2018.

\bibitem{guo2020deep}
Yulan Guo, Hanyun Wang, Qingyong Hu, Hao Liu, Li Liu, and Mohammed Bennamoun.
\newblock Deep learning for 3d point clouds: A survey.
\newblock {\em IEEE transactions on pattern analysis and machine intelligence},
  2020.

\bibitem{gupta2016cross}
Saurabh Gupta, Judy Hoffman, and Jitendra Malik.
\newblock Cross modal distillation for supervision transfer.
\newblock In {\em Proceedings of the IEEE conference on computer vision and
  pattern recognition}, pages 2827--2836, 2016.

\bibitem{hafiz2020survey}
Abdul~Mueed Hafiz and Ghulam~Mohiuddin Bhat.
\newblock A survey on instance segmentation: state of the art.
\newblock {\em International journal of multimedia information retrieval},
  pages 1--19, 2020.

\bibitem{han2020occuseg}
Lei Han, Tian Zheng, Lan Xu, and Lu Fang.
\newblock Occuseg: Occupancy-aware 3d instance segmentation.
\newblock In {\em Proceedings of the IEEE/CVF Conference on Computer Vision and
  Pattern Recognition}, pages 2940--2949, 2020.

\bibitem{hanocka2019meshcnn}
Rana Hanocka, Amir Hertz, Noa Fish, Raja Giryes, Shachar Fleishman, and Daniel
  Cohen-Or.
\newblock Meshcnn: a network with an edge.
\newblock {\em ACM Transactions on Graphics (TOG)}, 38(4):1--12, 2019.

\bibitem{he2021deep}
Yong He, Hongshan Yu, Xiaoyan Liu, Zhengeng Yang, Wei Sun, Yaonan Wang, Qiang
  Fu, Yanmei Zou, and Ajmal Mian.
\newblock Deep learning based 3d segmentation: A survey.
\newblock {\em arXiv preprint arXiv:2103.05423}, 2021.

\bibitem{Hedman21iccv_baking_NeRF}
Peter Hedman, Pratul Srinivasan, Ben Mildenhall, Jonathan Barron, and Paul
  Debevec.
\newblock {Baking Neural Radiance Fields for Real-Time View Synthesis}.
\newblock In {\em ICCV}, October 2021.

\bibitem{Henzler21cvpr_object_categories}
Philipp Henzler.
\newblock {Unsupervised Learning of 3D Object Categories from Videos in the
  Wild}.
\newblock In {\em CVPR}, 2021.

\bibitem{hermans2014dense}
Alexander Hermans, Georgios Floros, and Bastian Leibe.
\newblock Dense 3d semantic mapping of indoor scenes from rgb-d images.
\newblock In {\em 2014 IEEE International Conference on Robotics and Automation
  (ICRA)}, pages 2631--2638. IEEE, 2014.

\bibitem{jsenet}
Zeyu Hu, Mingmin Zhen, Xuyang Bai, Hongbo Fu, and Chiew-lan Tai.
\newblock Jsenet: Joint semantic segmentation and edge detection network for 3d
  point clouds.
\newblock In {\em ECCV}, 2020.

\bibitem{huang2019texturenet}
Jingwei Huang, Haotian Zhang, Li Yi, Thomas Funkhouser, Matthias Nie{\ss}ner,
  and Leonidas~J Guibas.
\newblock Texturenet: Consistent local parametrizations for learning from
  high-resolution signals on meshes.
\newblock In {\em Proceedings of the IEEE Conference on Computer Vision and
  Pattern Recognition}, pages 4440--4449, 2019.

\bibitem{huang2020lstm}
Rui Huang, Wanyue Zhang, Abhijit Kundu, Caroline Pantofaru, David~A Ross,
  Thomas Funkhouser, and Alireza Fathi.
\newblock An lstm approach to temporal 3d object detection in lidar point
  clouds.
\newblock In {\em Computer Vision--ECCV 2020: 16th European Conference,
  Glasgow, UK, August 23--28, 2020, Proceedings, Part XVIII 16}, pages
  266--282. Springer, 2020.

\bibitem{izadi2011kinectfusion}
Shahram Izadi, David Kim, Otmar Hilliges, David Molyneaux, Richard Newcombe,
  Pushmeet Kohli, Jamie Shotton, Steve Hodges, Dustin Freeman, Andrew Davison,
  et~al.
\newblock Kinectfusion: real-time 3d reconstruction and interaction using a
  moving depth camera.
\newblock In {\em Proceedings of the 24th annual ACM symposium on User
  interface software and technology}, pages 559--568, 2011.

\bibitem{Jain21iccv_DietNeRF}
Ajay Jain, Matthew Tancik, and Pieter Abbeel.
\newblock {Putting NeRF on a Diet: Semantically Consistent Few-Shot View
  Synthesis}.
\newblock In {\em ICCV}, October 2021.

\bibitem{Jang21iccv_CodeNeRF}
Wongbong Jang and Lourdes Agapito.
\newblock {CodeNeRF: Disentangled Neural Radiance Fields for Object
  Categories}.
\newblock In {\em ICCV}, October 2021.

\bibitem{jaritz2019multi}
Maximilian Jaritz, Jiayuan Gu, and Hao Su.
\newblock Multi-view pointnet for 3d scene understanding.
\newblock In {\em Proceedings of the IEEE International Conference on Computer
  Vision Workshops}, pages 0--0, 2019.

\bibitem{Jeong21iccv_SCNeRF}
Yoonwoo Jeong.
\newblock {Self-Calibrating Neural Radiance Fields}.
\newblock In {\em ICCV}, October 2021.

\bibitem{jin20203d}
Yiwei Jin, Diqiong Jiang, and Ming Cai.
\newblock 3d reconstruction using deep learning: a survey.
\newblock {\em Communications in Information and Systems}, 20(4):389--413,
  2020.

\bibitem{jing2020self}
Longlong Jing, Yucheng Chen, Ling Zhang, Mingyi He, and Yingli Tian.
\newblock Self-supervised modal and view invariant feature learning.
\newblock {\em arXiv preprint arXiv:2005.14169}, 2020.

\bibitem{johnson2017clevr}
Justin Johnson, Bharath Hariharan, Laurens Van Der~Maaten, Li Fei-Fei, C
  Lawrence~Zitnick, and Ross Girshick.
\newblock Clevr: A diagnostic dataset for compositional language and elementary
  visual reasoning.
\newblock In {\em Proceedings of the IEEE conference on computer vision and
  pattern recognition}, pages 2901--2910, 2017.

\bibitem{kingma2014adam}
Diederik~P Kingma and Jimmy Ba.
\newblock Adam: A method for stochastic optimization.
\newblock {\em arXiv preprint arXiv:1412.6980}, 2014.

\bibitem{koepke2020sight}
A~Sophia Koepke, Olivia Wiles, Yael Moses, and Andrew Zisserman.
\newblock Sight to sound: An end-to-end approach for visual piano
  transcription.
\newblock In {\em ICASSP 2020-2020 IEEE International Conference on Acoustics,
  Speech and Signal Processing (ICASSP)}, pages 1838--1842. IEEE, 2020.

\bibitem{korbar2018cooperative}
Bruno Korbar, Du Tran, and Lorenzo Torresani.
\newblock Cooperative learning of audio and video models from self-supervised
  synchronization.
\newblock In {\em Advances in Neural Information Processing Systems}, pages
  7763--7774, 2018.

\bibitem{Kosiorek21icml_NeRF_VAE}
Adam Kosiorek, Heiko Strathmann, Daniel Zoran, Pol Moreno, Rosalia Schneider,
  Sona Mokrá, and Danilo Rezende.
\newblock {NeRF-VAE: A Geometry Aware 3D Scene Generative Model}.
\newblock In {\em ICML}, 2021.

\bibitem{kowdle2018need4speed}
Adarsh Kowdle, Christoph Rhemann, Sean Fanello, Andrea Tagliasacchi, Jonathan
  Taylor, Philip Davidson, Mingsong Dou, Kaiwen Guo, Cem Keskin, Sameh Khamis,
  et~al.
\newblock The need 4 speed in real-time dense visual tracking.
\newblock {\em ACM Transactions on Graphics (TOG)}, 37(6):1--14, 2018.

\bibitem{krizhevsky2012imagenet}
Alex Krizhevsky, Ilya Sutskever, and Geoffrey~E Hinton.
\newblock Imagenet classification with deep convolutional neural networks.
\newblock {\em Advances in neural information processing systems},
  25:1097--1105, 2012.

\bibitem{kundu2020virtual}
Abhijit Kundu, Xiaoqi Yin, Alireza Fathi, David Ross, Brian Brewington, Thomas
  Funkhouser, and Caroline Pantofaru.
\newblock Virtual multi-view fusion for 3d semantic segmentation.
\newblock In {\em European Conference on Computer Vision}, pages 518--535.
  Springer, 2020.

\bibitem{6907298}
K. {Lai}, L. {Bo}, and D. {Fox}.
\newblock Unsupervised feature learning for 3d scene labeling.
\newblock In {\em 2014 IEEE International Conference on Robotics and Automation
  (ICRA)}, pages 3050--3057, 2014.

\bibitem{lateef2019survey}
Fahad Lateef and Yassine Ruichek.
\newblock Survey on semantic segmentation using deep learning techniques.
\newblock {\em Neurocomputing}, 338:321--348, 2019.

\bibitem{lawin2017deep}
Felix~J{\"a}remo Lawin, Martin Danelljan, Patrik Tosteberg, Goutam Bhat,
  Fahad~Shahbaz Khan, and Michael Felsberg.
\newblock Deep projective 3d semantic segmentation.
\newblock In {\em International Conference on Computer Analysis of Images and
  Patterns}, pages 95--107. Springer, 2017.

\bibitem{Lin21iccv_BARF}
Chen-Hsuan Lin, Wei-Chiu Ma, Antonio Torralba, and Simon Lucey.
\newblock {BARF: Bundle-Adjusting Neural Radiance Fields}.
\newblock In {\em ICCV}, October 2021.

\bibitem{lin2014mscoco}
Tsung-Yi Lin, Michael Maire, Serge Belongie, James Hays, Pietro Perona, Deva
  Ramanan, Piotr Doll{\'a}r, and C~Lawrence Zitnick.
\newblock Microsoft coco: Common objects in context.
\newblock In {\em European conference on computer vision}, pages 740--755.
  Springer, 2014.

\bibitem{Lindell20cvpr_AutoInt}
David Lindell, Julien Martel, and Gordon Wetzstein.
\newblock {AutoInt: Automatic Integration for Fast Neural Volume Rendering}.
\newblock In {\em CVPR}, 2021.

\bibitem{Liu20neurips_sparse_nerf}
Lingjie Liu, Jiatao Gu, Kyaw~Zaw Lin, Tat-Seng Sua, and Christian Theobalt.
\newblock {Neural Sparse Voxel Fields}.
\newblock In {\em Adv. Neural Inform. Process. Syst.}, 2020.

\bibitem{liu2020deep}
Li Liu, Wanli Ouyang, Xiaogang Wang, Paul Fieguth, Jie Chen, Xinwang Liu, and
  Matti Pietik{\"a}inen.
\newblock Deep learning for generic object detection: A survey.
\newblock {\em International journal of computer vision}, 128(2):261--318,
  2020.

\bibitem{liu2019deep}
Weiping Liu, Jia Sun, Wanyi Li, Ting Hu, and Peng Wang.
\newblock Deep learning on point clouds and its application: A survey.
\newblock {\em Sensors}, 19(19):4188, 2019.

\bibitem{long2015fully}
Jonathan Long, Evan Shelhamer, and Trevor Darrell.
\newblock Fully convolutional networks for semantic segmentation.
\newblock In {\em Proceedings of the IEEE conference on computer vision and
  pattern recognition}, pages 3431--3440, 2015.

\bibitem{ma2017multi}
Lingni Ma, J{\"o}rg St{\"u}ckler, Christian Kerl, and Daniel Cremers.
\newblock Multi-view deep learning for consistent semantic mapping with rgb-d
  cameras.
\newblock In {\em 2017 IEEE/RSJ International Conference on Intelligent Robots
  and Systems (IROS)}, pages 598--605. IEEE, 2017.

\bibitem{MartinBrualla21cvpr_nerfw}
Ricardo Martin-Brualla, Noha Radwan, Mehdi Sajjadi, Jonathan Barron, Alexey
  Dosovitskiy, and Daniel Duckworth.
\newblock {NeRF in the Wild: Neural Radiance Fields for Unconstrained Photo
  Collections}.
\newblock In {\em CVPR}, 2021.

\bibitem{mascaro2021diffuser}
Ruben Mascaro, Lucas Teixeira, and Margarita Chli.
\newblock Diffuser: Multi-view 2d-to-3d label diffusion for semantic scene
  segmentation.
\newblock In {\em IEEE International Conference on Robotics and Automation
  (ICRA 2021)(virtual)}, 2021.

\bibitem{masci2015geodesic}
Jonathan Masci, Davide Boscaini, Michael Bronstein, and Pierre Vandergheynst.
\newblock Geodesic convolutional neural networks on riemannian manifolds.
\newblock In {\em Proceedings of the IEEE international conference on computer
  vision workshops}, pages 37--45, 2015.

\bibitem{Max1995tvcg_VolumeRendering}
Nelson Max.
\newblock Optical models for direct volume rendering.
\newblock {\em IEEE Transactions on Visualization and Computer Graphics},
  1(2):99--108, 1995.

\bibitem{mccormac2017semanticfusion}
John McCormac, Ankur Handa, Andrew Davison, and Stefan Leutenegger.
\newblock Semanticfusion: Dense 3d semantic mapping with convolutional neural
  networks.
\newblock In {\em 2017 IEEE International Conference on Robotics and automation
  (ICRA)}, pages 4628--4635. IEEE, 2017.

\bibitem{Meng21iccv_GNeRF}
Quan Meng.
\newblock {GNeRF: GAN-Based Neural Radiance Field Without Posed Camera}.
\newblock In {\em ICCV}, October 2021.

\bibitem{mescheder2019occupancy}
Lars Mescheder, Michael Oechsle, Michael Niemeyer, Sebastian Nowozin, and
  Andreas Geiger.
\newblock Occupancy networks: Learning 3d reconstruction in function space.
\newblock In {\em Proceedings of the IEEE/CVF Conference on Computer Vision and
  Pattern Recognition}, pages 4460--4470, 2019.

\bibitem{Mescheder19cvpr_occupancy_net}
Lars Mescheder, Michael Oechsle, Michael Niemeyer, Sebastian Nowozin, and
  Andreas Geiger.
\newblock {Occupancy Networks: Learning 3D Reconstruction in Function Space}.
\newblock In {\em CVPR}, 2019.

\bibitem{meyerimproving}
Johannes Meyer, Andreas Eitel, Thomas Brox, and Wolfram Burgard.
\newblock Improving unimodal object recognition with multimodal contrastive
  learning.
\newblock In {\em IEEE/RSJ International Conference on Intelligent Robots and
  Systems}, 2020.

\bibitem{Mildenhall20eccv_nerf}
Ben Mildenhall, Pratul Srinivasan, Matthew Tancik, Jonathan Barron, Ravi
  Ramamoorthi, and Ren Ng.
\newblock {NeRF: Representing Scenes as Neural Radiance Fields for View
  Synthesis}.
\newblock In {\em ECCV}, pages 405--421. Springer, 2020.

\bibitem{milioto2019iros}
A. Milioto, I. Vizzo, J. Behley, and C. Stachniss.
\newblock {RangeNet++: Fast and Accurate LiDAR Semantic Segmentation}.
\newblock In {\em IEEE/RSJ Intl.~Conf.~on Intelligent Robots and Systems
  (IROS)}, 2019.

\bibitem{minaee2021image}
Shervin Minaee, Yuri~Y Boykov, Fatih Porikli, Antonio~J Plaza, Nasser
  Kehtarnavaz, and Demetri Terzopoulos.
\newblock Image segmentation using deep learning: A survey.
\newblock {\em IEEE Transactions on Pattern Analysis and Machine Intelligence},
  2021.

\bibitem{Murez20eccv_atlas}
Zak Murez, Tarrence van As, James Bartolozzi, Ayan Sinha, Vijay Badrinarayanan,
  and Andrew Rabinovich.
\newblock Atlas: End-to-end 3d scene reconstruction from posed images.
\newblock In {\em ECCV}, 2020.

\bibitem{nagrani2020speech2action}
Arsha Nagrani, Chen Sun, David Ross, Rahul Sukthankar, Cordelia Schmid, and
  Andrew Zisserman.
\newblock Speech2action: Cross-modal supervision for action recognition.
\newblock In {\em Proceedings of the IEEE Conference on Computer Vision and
  Pattern Recognition}, 2020.

\bibitem{Neff21egsr_DONeRF}
Thomas Neff.
\newblock {DONeRF: Towards Real-Time Rendering of Compact Neural Radiance
  Fields using Depth Oracle Networks}.
\newblock In {\em Eurographics}, 2021.

\bibitem{Niemeyer21cvpr_GIRAFFE}
Michael Niemeyer and Andreas Geiger.
\newblock {GIRAFFE: Representing Scenes as Compositional Generative Neural
  Feature Fields}.
\newblock In {\em CVPR}, 2021.

\bibitem{Niemeyer20cvpr_DVR}
Michael Niemeyer, Lars Mescheder, Michael Oechsle, and Andreas Geiger.
\newblock {Differentiable Volumetric Rendering: Learning Implicit 3D
  Representations without 3D Supervision}.
\newblock In {\em CVPR}, 2020.

\bibitem{Noguchi21iccv_NARF}
Atsuhiro Noguchi.
\newblock {Neural Articulated Radiance Field}.
\newblock In {\em ICCV}, October 2021.

\bibitem{Oechsle21iccv_UNISURF}
Michael Oechsle, Songyou Peng, and Andreas Geiger.
\newblock {UNISURF: Unifying Neural Implicit Surfaces and Radiance Fields for
  Multi-View Reconstruction}.
\newblock In {\em ICCV}, October 2021.

\bibitem{owens2016ambient}
Andrew Owens, Jiajun Wu, Josh~H. McDermott, William~T. Freeman, and Antonio
  Torralba.
\newblock Ambient sound provides supervision for visual learning.
\newblock In {\em Proceedings of the European Conference on Computer Vision
  (ECCV)}, 2016.

\bibitem{park2019deepsdf}
Jeong~Joon Park, Peter Florence, Julian Straub, Richard Newcombe, and Steven
  Lovegrove.
\newblock Deepsdf: Learning continuous signed distance functions for shape
  representation.
\newblock In {\em Proceedings of the IEEE/CVF Conference on Computer Vision and
  Pattern Recognition}, pages 165--174, 2019.

\bibitem{Park19cvpr_deepsdf}
Jeong~Joon Park, Pete Florence, Julian Straub, Richard Newcombe, and Steven
  Lovegrove.
\newblock {DeepSDF: Learning Continuous Signed Distance Functions for Shape
  Representation}.
\newblock In {\em CVPR}, 2019.

\bibitem{Park21iccv_nerfies}
Keunhong Park, Utkarsh Sinha, Jonathan Barron, Sofien Bouaziz, Dan Goldman,
  Steven Seitz, and Ricardo Martin-Brualla.
\newblock {Nerfies: Deformable Neural Radiance Fields}.
\newblock In {\em ICCV}, October 2021.

\bibitem{Peng21cvpr_NeuralBody}
Sida Peng.
\newblock {Neural body: Implicit neural representations with structured latent
  codes for novel view synthesis of dynamic humans.}
\newblock In {\em CVPR}, 2021.

\bibitem{Peng2020ECCV}
Songyou Peng, Michael Niemeyer, Lars Mescheder, Marc Pollefeys, and Andreas
  Geiger.
\newblock Convolutional occupancy networks.
\newblock In {\em European Conference on Computer Vision (ECCV)}, Cham, Aug.
  2020. Springer International Publishing.

\bibitem{pham2019jsis3d}
Quang-Hieu Pham, Thanh Nguyen, Binh-Son Hua, Gemma Roig, and Sai-Kit Yeung.
\newblock Jsis3d: joint semantic-instance segmentation of 3d point clouds with
  multi-task pointwise networks and multi-value conditional random fields.
\newblock In {\em Proceedings of the IEEE Conference on Computer Vision and
  Pattern Recognition}, pages 8827--8836, 2019.

\bibitem{Chibane20cvpr_ifnet}
Gerard Pons-Moll.
\newblock {Implicit Functions in Feature Space for 3D Shape Reconstruction and
  Completion}.
\newblock In {\em CVPR}, 2020.

\bibitem{qi2017pointnet}
Charles~R Qi, Hao Su, Kaichun Mo, and Leonidas~J Guibas.
\newblock Pointnet: Deep learning on point sets for 3d classification and
  segmentation.
\newblock In {\em Proceedings of the IEEE conference on computer vision and
  pattern recognition}, pages 652--660, 2017.

\bibitem{qi2017pointnet++}
Charles~Ruizhongtai Qi, Li Yi, Hao Su, and Leonidas~J Guibas.
\newblock Pointnet++: Deep hierarchical feature learning on point sets in a
  metric space.
\newblock In {\em Advances in neural information processing systems}, pages
  5099--5108, 2017.

\bibitem{Rebain20arxiv_derf}
Daniel Rebain, Wei Jiang, Soroosh Yazdani, Ke Li, Kwang~Moo Yi, and Andrea
  Tagliasacchi.
\newblock {DeRF: Decomposed Radiance Fields}.
\newblock {\em https://arxiv.org/abs/2011.12490}, 2020.

\bibitem{Reiser21iccv_KiloNeRF}
Christian Reiser, Songyou Peng, Yiyi Liao, and Andreas Geiger.
\newblock {KiloNeRF: Speeding Up Neural Radiance Fields With Thousands of Tiny
  MLPs}.
\newblock In {\em ICCV}, October 2021.

\bibitem{Rematas21icml_ShaRF}
Konstantinos Rematas, Ricardo Martin-Brualla, and Vittorio Ferrari.
\newblock {ShaRF: Shape-conditioned Radiance Fields from a Single View}.
\newblock In {\em ICML}, 2021.

\bibitem{riegler2017octnet}
Gernot Riegler, Ali Osman~Ulusoy, and Andreas Geiger.
\newblock Octnet: Learning deep 3d representations at high resolutions.
\newblock In {\em Proceedings of the IEEE Conference on Computer Vision and
  Pattern Recognition}, pages 3577--3586, 2017.

\bibitem{ronneberger2015unet}
Olaf Ronneberger, Philipp Fischer, and Thomas Brox.
\newblock U-net: Convolutional networks for biomedical image segmentation.
\newblock In {\em International Conference on Medical image computing and
  computer-assisted intervention}, pages 234--241. Springer, 2015.

\bibitem{ronneberger2015u}
Olaf Ronneberger, Philipp Fischer, and Thomas Brox.
\newblock U-net: Convolutional networks for biomedical image segmentation.
\newblock In {\em International Conference on Medical image computing and
  computer-assisted intervention}, pages 234--241. Springer, 2015.

\bibitem{Saito19Iccv_PIFu}
Shunsuke Saito, Zeng Huang, Ryota Natsume, Shigeo Morishima, Angjoo Kanazawa,
  and Hao Li.
\newblock {PIFu: Pixel-Aligned Implicit Function for High-Resolution Clothed
  Human Digitization}.
\newblock In {\em ICCV}, October 2019.

\bibitem{Schwarz20neurips_graf}
Katja Schwarz, Yiyi Liao, Michael Niemeyer, and Andreas Geiger.
\newblock {GRAF: Generative Radiance Fields for 3D-Aware Image Synthesis}.
\newblock In {\em Adv. Neural Inform. Process. Syst.}, 2020.

\bibitem{shi2019pv}
Shaoshuai Shi, Chaoxu Guo, Li Jiang, Zhe Wang, Jianping Shi, Xiaogang Wang, and
  Hongsheng Li.
\newblock {PV-RCNN}: Point-voxel feature set abstraction for 3d object
  detection.
\newblock In {\em Proceedings of the IEEE/CVF Conference on Computer Vision and
  Pattern Recognition}, pages 10529--10538, 2020.

\bibitem{song2017semantic}
Shuran Song, Fisher Yu, Andy Zeng, Angel~X Chang, Manolis Savva, and Thomas
  Funkhouser.
\newblock Semantic scene completion from a single depth image.
\newblock In {\em Proceedings of the IEEE Conference on Computer Vision and
  Pattern Recognition}, pages 1746--1754, 2017.

\bibitem{Srinivasan21cvpr_NeRV}
Pratul Srinivasan, Boyang Deng, Xiuming Zhang, Matthew Tancik, Ben Mildenhall,
  and Jonathan Barron.
\newblock {NeRV: Neural Reflectance and Visibility Fields for Relighting and
  View Synthesis}.
\newblock In {\em CVPR}, 2021.

\bibitem{straub2019replica}
Julian Straub, Thomas Whelan, Lingni Ma, Yufan Chen, Erik Wijmans, Simon Green,
  Jakob~J Engel, Raul Mur-Artal, Carl Ren, Shobhit Verma, et~al.
\newblock The replica dataset: A digital replica of indoor spaces.
\newblock {\em arXiv preprint arXiv:1906.05797}, 2019.

\bibitem{Sucar21iccv_iMAP}
Edgar Sucar, Shikun Liu, Joseph Ortiz, and Andrew Davison.
\newblock {iMAP: Implicit Mapping and Positioning in Real-Time}.
\newblock In {\em ICCV}, October 2021.

\bibitem{sun2020scalability}
Pei Sun, Henrik Kretzschmar, Xerxes Dotiwalla, Aurelien Chouard, Vijaysai
  Patnaik, Paul Tsui, James Guo, Yin Zhou, Yuning Chai, Benjamin Caine, et~al.
\newblock Scalability in perception for autonomous driving: Waymo open dataset.
\newblock In {\em Proceedings of the IEEE/CVF Conference on Computer Vision and
  Pattern Recognition}, pages 2446--2454, 2020.

\bibitem{Takikawa21cvpr_nglod}
Towaki Takikawa.
\newblock {Neural geometric level of detail: Real-time rendering with implicit
  3d shapes.}
\newblock In {\em CVPR}, 2021.

\bibitem{thomas2019kpconv}
Hugues Thomas, Charles~R Qi, Jean-Emmanuel Deschaud, Beatriz Marcotegui,
  Fran{\c{c}}ois Goulette, and Leonidas~J Guibas.
\newblock Kpconv: Flexible and deformable convolution for point clouds.
\newblock In {\em Proceedings of the IEEE International Conference on Computer
  Vision}, pages 6411--6420, 2019.

\bibitem{tian2019contrastive}
Yonglong Tian, Dilip Krishnan, and Phillip Isola.
\newblock Contrastive multiview coding.
\newblock In {\em Computer Vision--ECCV 2020: 16th European Conference,
  Glasgow, UK, August 23--28, 2020, Proceedings, Part XI 16}, pages 776--794.
  Springer, 2020.

\bibitem{tran20153dconvnets}
Du Tran, Lubomir Bourdev, Rob Fergus, Lorenzo Torresani, and Manohar Paluri.
\newblock Learning spatiotemporal features with 3d convolutional networks.
\newblock In {\em Proceedings of the IEEE international conference on computer
  vision}, pages 4489--4497, 2015.

\bibitem{Tretschk21iccv_NR_NeRF}
Edgar Tretschk, Ayush Tewari, Vladislav Golyanik, Michael Zollhöfer, Christoph
  Lassner, and Christian Theobalt.
\newblock {Non-Rigid Neural Radiance Fields: Reconstruction and Novel View
  Synthesis of a Deforming Scene from Monocular Video}.
\newblock In {\em ICCV}, October 2021.

\bibitem{vineet2015incremental}
Vibhav Vineet, Ondrej Miksik, Morten Lidegaard, Matthias Nie{\ss}ner, Stuart
  Golodetz, Victor~A Prisacariu, Olaf K{\"a}hler, David~W Murray, Shahram
  Izadi, Patrick P{\'e}rez, et~al.
\newblock Incremental dense semantic stereo fusion for large-scale semantic
  scene reconstruction.
\newblock In {\em 2015 IEEE International Conference on Robotics and Automation
  (ICRA)}, pages 75--82. IEEE, 2015.

\bibitem{vora2020pointpainting}
Sourabh Vora, Alex~H Lang, Bassam Helou, and Oscar Beijbom.
\newblock Pointpainting: Sequential fusion for 3d object detection.
\newblock In {\em Proceedings of the IEEE/CVF Conference on Computer Vision and
  Pattern Recognition}, pages 4604--4612, 2020.

\bibitem{wang2019towards}
Haiyan Wang, Xuejian Rong, Liang Yang, Shuihua Wang, and Yingli Tian.
\newblock Towards weakly supervised semantic segmentation in 3{D}
  graph-structured point clouds of wild scenes.
\newblock In {\em BMVC}, page 284, 2019.

\bibitem{Wang21cvpr_IBRNet}
Qianqian Wang, Zhicheng Wang, Kyle Genova, Pratul Srinivasan, Howard Zhou,
  Jonathan Barron, Ricardo Martin-Brualla, Noah Snavely, and Thomas Funkhouser.
\newblock {IBRNet: Learning Multi-View Image-Based Rendering}.
\newblock In {\em CVPR}, 2021.

\bibitem{Wang21arxiv_NeRFminusminus}
Zirui Wang, Shangzhe Wu, Weidi Xie, Min Chen, and Victor~Adrian Prisacariu.
\newblock {NeRF--: Neural Radiance Fields Without Known Camera Parameters}.
\newblock {\em https://arxiv.org/abs/2102.07064}, 2021.

\bibitem{wu2019wider}
Zifeng Wu, Chunhua Shen, and Anton Van Den~Hengel.
\newblock Wider or deeper: Revisiting the resnet model for visual recognition.
\newblock {\em Pattern Recognition}, 90:119--133, 2019.

\bibitem{Yariv20neurips_MVNeuralSurfaceRecon}
Lior Yariv, Yoni Kasten, Dror Moran, Meirav Galun, Matan Atzmon, Ronan Basri,
  and Yaron Lipman.
\newblock {Multiview Neural Surface Reconstruction by Disentangling Geometry
  and Appearance}.
\newblock In {\em Adv. Neural Inform. Process. Syst.}, 2020.

\bibitem{YenChen20iros_iNeRF}
Lin Yen-Chen, Pete Florence, Jonathan Barron, Alberto Rodriguez, Phillip Isola,
  and Tsung-Yi Lin.
\newblock {iNeRF: Inverting Neural Radiance Fields for Pose Estimation}.
\newblock In {\em IROS}, 2021.

\bibitem{Yu21iccv_PlenOctrees}
Alex Yu, Ruilong Li, Matthew Tancik, Hao Li, Ren Ng, and Angjoo Kanazawa.
\newblock {PlenOctrees for Real-Time Rendering of Neural Radiance Fields}.
\newblock In {\em ICCV}, October 2021.

\bibitem{Yu21cvpr_pixelNeRF}
Alex Yu, Vickie Ye, Matthew Tancik, and Angjoo Kanazawa.
\newblock {pixelNeRF: Neural Radiance Fields from One or Few Images}.
\newblock In {\em CVPR}, 2021.

\bibitem{Yuan21cvpr_STaR}
Wentao Yuan, Zhaoyang Lv, Tanner Schmidt, and Steven Lovegrove.
\newblock {STaR: Self-supervised Tracking and Reconstruction of Rigid Objects
  in Motion with Neural Rendering}.
\newblock In {\em CVPR}, 2021.

\bibitem{hdrihaven}
Greg Zaal, Rob Tuytel, Rico Cilliers, James~Ray Cock, Andreas Mischok, Sergej
  Majboroda, Dimitrios Savva, and Jurita Burger.
\newblock Hdri haven.
\newblock \url{hhttps://polyhaven.com/hdris}, 2021.

\bibitem{zhang2019large}
Cheng Zhang, Zhi Liu, Guangwen Liu, and Dandan Huang.
\newblock Large-scale 3d semantic mapping using monocular vision.
\newblock In {\em 2019 IEEE 4th International Conference on Image, Vision and
  Computing (ICIVC)}, pages 71--76. IEEE, 2019.

\bibitem{zhang2021nerfactor}
Xiuming Zhang, Pratul~P Srinivasan, Boyang Deng, Paul Debevec, William~T
  Freeman, and Jonathan~T Barron.
\newblock Nerfactor: Neural factorization of shape and reflectance under an
  unknown illumination.
\newblock {\em arXiv preprint arXiv:2106.01970}, 2021.

\bibitem{zhang2020polarnet}
Yang Zhang, Zixiang Zhou, Philip David, Xiangyu Yue, Zerong Xi, Boqing Gong,
  and Hassan Foroosh.
\newblock Polarnet: An improved grid representation for online lidar point
  clouds semantic segmentation.
\newblock In {\em Proceedings of the IEEE/CVF Conference on Computer Vision and
  Pattern Recognition}, pages 9601--9610, 2020.

\bibitem{zhao2018soundofpixels}
Hang Zhao, Chuang Gan, Andrew Rouditchenko, Carl Vondrick, Josh McDermott, and
  Antonio Torralba.
\newblock The sound of pixels.
\newblock In {\em Proceedings of the European Conference on Computer Vision
  (ECCV)}, page 570–586, 2018.

\bibitem{Zhi21iccv_SemanticNeRF}
Shuaifeng Zhi, Tristan Laidlow, Stefan Leutenegger, and Andrew Davison.
\newblock {In-Place Scene Labelling and Understanding with Implicit Scene
  Representation}.
\newblock In {\em ICCV}, October 2021.

\bibitem{zhu2020cylindrical}
Xinge Zhu, Hui Zhou, Tai Wang, Fangzhou Hong, Yuexin Ma, Wei Li, Hongsheng Li,
  and Dahua Lin.
\newblock Cylindrical and asymmetrical 3d convolution networks for lidar
  segmentation.
\newblock In {\em Proceedings of the IEEE/CVF Conference on Computer Vision and
  Pattern Recognition}, pages 9939--9948, 2021.

\end{thebibliography}
}

\appendix

\twocolumn[
\centering
\Large
\textbf{NeSF: Neural Semantic Fields for \\ Generalizable Semantic Segmentation of 3D Scenes} \\
\vspace{0.5em}Supplementary Material \\
\vspace{1.0em}
] %

\begin{figure}
\begin{center}
    \includegraphics[width=.95\linewidth]{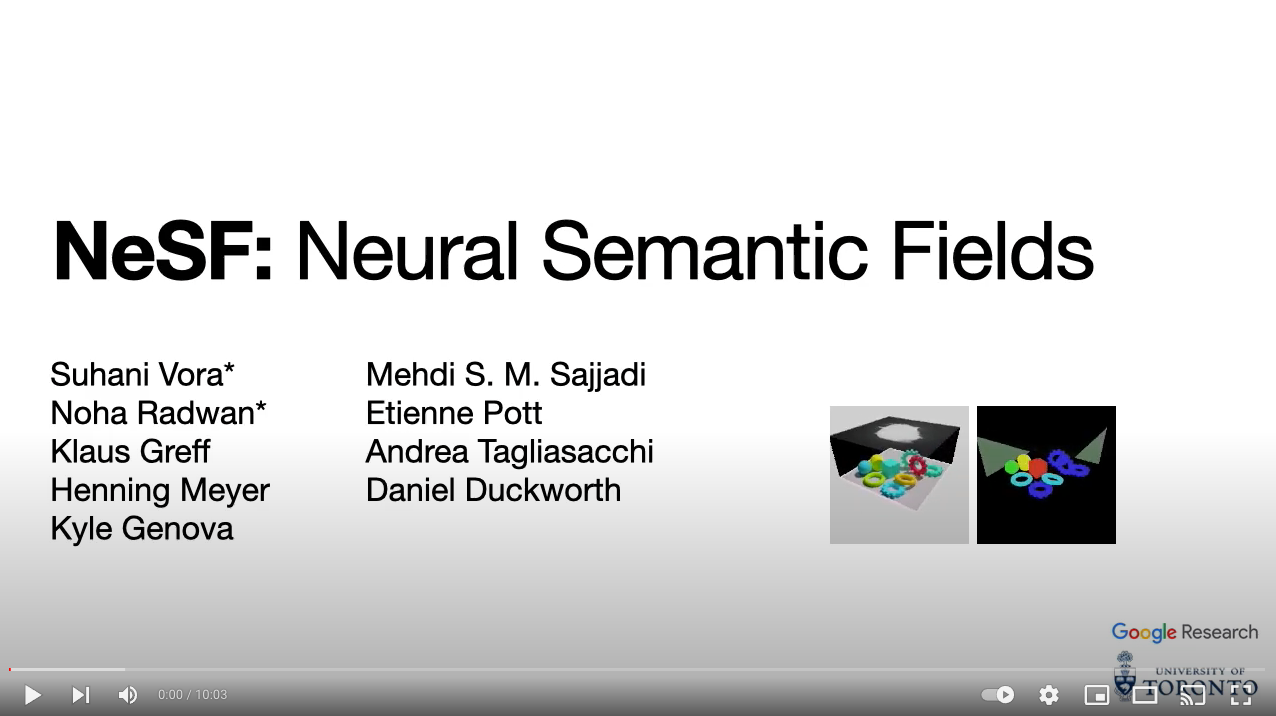}
\end{center}
\vspace{-1em}
\caption{
    \textbf{Overview Video -- }
    We highly recommend viewing the high-definition overview video on \href{\YouTubeLink}{YouTube}. % and full-resolution animations in the ZIP file accompanying this PDF.
}
\label{fig:overview_view}
\end{figure}
\section{Contributions}

We describe the specific contributions of each individual author in the following.

\bigname{Suhani Vora} was one of two first authors on this work.
She implemented \NeSF{}'s 2D inference framework; implemented and ran \NeSF{} experiments; implemented and ran \DeepLab{} baseline experiments; setup the project website; and wrote portions of the paper.

\bigname{Noha Radwan} was one of two first authors on this work.
She implemented the 3D semantic reasoning module; implemented and ran \NeSF{} experiments; implemented \NeSF{}'s 2D evaluation framework; setup the project website; and edited the paper.

\bigname{Klaus Greff} was responsible for Kubric~\cite{greff2021kubric}, the technology used to generate the datasets in this work.
He further provided assistance in development of the datasets and the animations in the overview video.

\bigname{Henning Meyer} assisted in the development of the \ToyBoxEasy{} and \ToyBoxHard{} datasets.
He further contributed significantly to the codebase upon which \NeSF{} was built.

\bigname{Kyle Genova} was responsible for the \SparseConvNet{} baseline.
He further assisted in experiment design and wrote portions of the paper.

\bigname{Mehdi S. M. Sajjadi} suggested experiments and contributed to the codebase upon which \NeSF{} was built

\bigname{Etienne Pot} implemented the scalable dataset pipeline used to train \NeSF{}.
He further contributed significantly to the codebase upon which \NeSF{} was built.

\bigname{Andrea Tagliasacchi} oversaw the project as a research lead.
He proposed the initial \NeSF{} model architecture 
He further suggested experiments; and oversaw and wrote large portions of the paper.

\bigname{Daniel Duckworth} oversaw the project as a technical lead.
He coordinated contributors and designed and prototyped the software architecture within which \NeSF{} was implemented.
He further suggested, implemented, and ran \NeSF{} experiments; generated the datasets used in this paper; implemented \NeSF{}'s 3D inference framework; wrote portions of the paper; generated visualizations; and wrote, assembled, and recorded the overview video.
\section{Training Details}

We describe the model architecture and training procedure used by \NeSF{} and its baselines below.
Unless otherwise stated, we train each method using all \TrainScenes{} with \NumTrainImagesPerScene{} randomly-selected images per scene.
For \KLEVR{}, this results in \NumKLEVRTrainScenes{} \TrainScenes{}; for \ToyBoxEasy{} and \NumToyBoxTrainScenes{} \TrainScenes{} for  \ToyBoxHard{}.
All methods are evaluated on \NumTestImagesPerScene{} randomly-selected images from each dataset's \TestScenes{}.
For \KLEVR{}, this results in \NumKLEVRTestScenes{} \TestScenes{}; for \ToyBoxEasy{} and \ToyBoxHard{}, \NumToyBoxTestScenes{} \TestScenes{} each.
We ensure that each method observes the same randomly-selected set of images per scene by specifying the seed of the random number generator.

\paragraph{\NeRF{}}
The first stage of \NeSF{} is the training of per-scene \NeRF{} models.
We employ the model architecture and training regime of Mildenhall~\textit{et al.}~\cite{Mildenhall20eccv_nerf}.
Each scene's density field is described by an MLP with 8 hidden layers of 256 units, and its appearance by an additional MLP of 1 hidden layer and 128 units.
We employ 10 octaves for positional encoding.
Each \NeRF{} model is trained on pixels selected at random from \NumTrainImagesPerScene{} views with the Adam optimizer~\cite{kingma2014adam}.
The learning rate is exponentially decayed from $1e{-}3$ to $5.4e{-}4$ over 25,000 steps.
We train each \NeRF{} model for approximately 20 minutes on 8 TPUv2 cores.

\paragraph{\NeSF{}} 
\NeSF{} has two major model components: a 3D UNet and an MLP Decoder.
For the 3D UNet, we employ the UNet architecture of {\c{C}}i{\c{c}}ek~\textit{et al.}~\cite{cciccek20163d} with the BatchNorm layers removed and only 2 max-pooling operations.
We use 32, 64, and 128 output channels prior to each max-pooling operation.
For the MLP Decoder, we employ 2 hidden layers of 128 hidden units each with a ReLU non-linearity.

We train \NeSF{} with Adam optimizer~\cite{kingma2014adam}.
We use an exponentially decaying learning rate initialized to $1e{-}3$ and decaying to $1e{-}5$ over 25,000 steps.
At each step, we employ a stratified sampling approach: we randomly select 32 scenes, then randomly select a set of 128 pixels from each scene's \TrainViews{}.
For volumetric rendering, we sample 192 points along each ray according to the stratified approach employed in \NeRF{}~\cite{Mildenhall20eccv_nerf}.
For each scene in the batch, we discretize \NeRF{}'s density grid by probing at $64^3$ evenly-spaced points.
Before discretizing, we apply a random rotation about the z-axis (upwards) to each scene.
For smoothness regularization, we uniformly sample 8,192 additional 3D coordinates from each scene
and add random noise with standard deviation 0.05.
When computing the loss, we assign a weight of 0.1 to the smoothness regularization term.

We find that we are able to train \NeSF{} to convergence in approximately 45 minutes on 32 TPUv3 cores.

\paragraph{\DeepLab{}} 
We train a DeepLab Wide-ResNet-38 model \cite{wu2019wider}, warm starting with a checkpoint pre-trained on COCO. 
For our optimization scheme, we apply SGD + Momentum with a slow start learning rate of 1e-4 and a linear ramp up to 6e-3 followed by a cosine schedule decay in learning rate to 1.26e-7 at 55,000 training steps. 
We additionally apply weight decay of $1.0e{-}4$. 
For each train step, we use a batch size of 32. 
Models are trained on 32 TPUv3 chips. 
To enable re-use of a well-performing hyperparameter configuration, we up-sample our input images from 256x256 to 1024x1024, using bilinear interpolation for the RGB input and nearest neighbor interpolation for the corresponding semantic maps. 

\paragraph{\SparseConvNet{}}
Our \SparseConvNet{}~\cite{graham20183d} implementation is based on the TF3D~\cite{huang2020lstm} and 2D3DNet~\cite{genova2021learning} implementations. 
Each convolutional layer except the last is an occupancy-normalized 3x3x3 sparse spatial convolution followed by batch norm and then ReLU. 
The final layer omits batch norm and ReLU. 
Each encoder stage is a pair of convolution layers followed by a 2x2x2 spatial max-pool operation, and each decoder layer is a voxel unpooling operation followed by a pair of convolutional layers. 
The encoder feature widths are (64, 64), (64, 96), (96, 128), (128, 160), (160, 192), (192, 224), (224, 256). 
These are the output channel counts of the first and second convolutional layers per block. 
The bottleneck is a sequence of two convolutional layers of widths 256 each. 
The decoder feature widths are (256, 256), (224, 224), (192, 192), (160, 160), (128, 128), (96, 96), (64, 64). 
Finally we apply a sequence of three convolutional layers with sizes (64, 64, \texttt{class\_count}), which are followed by a softmax layer and a cross-entropy loss function. 
Our input features are only occupancy (i.e., a 1 on all input points). 
We use 0.005-width voxels in a [-1, 1] cube scene. 
We optimize for 450,000 steps with SGD using a momentum of 0.9, a batch size of 5, an initial learning rate of 0.015, and a cosine learning rate decay starting at step 200,000 and ending at step 450,000. 
We add an $\ell_2$ weight decay loss of $1e{-}4$ and train asynchronously on 20 NVIDIA V100 GPUs. 
We apply the following data augmentations: XY rotations of up to $\pm 10$ degrees, z rotations of $\pm 180$ degrees, and a random scale factor between 0.9 and 1.1.
\section{Analysis}

\subsection{Qualitative Results}

% Figure from this slide,
% https://colab.corp.google.com/drive/1Ac4oCld4E8uCAJYDpw4qgdZpeRCU3qGV?resourcekey=0-na0gkrIDeli7yDUv1EILqw#scrollTo=vcbf8Ur-zQa7&line=18&uniqifier=1

\begin{figure*}[t]
\centering
    \includegraphics[width=0.95\linewidth]{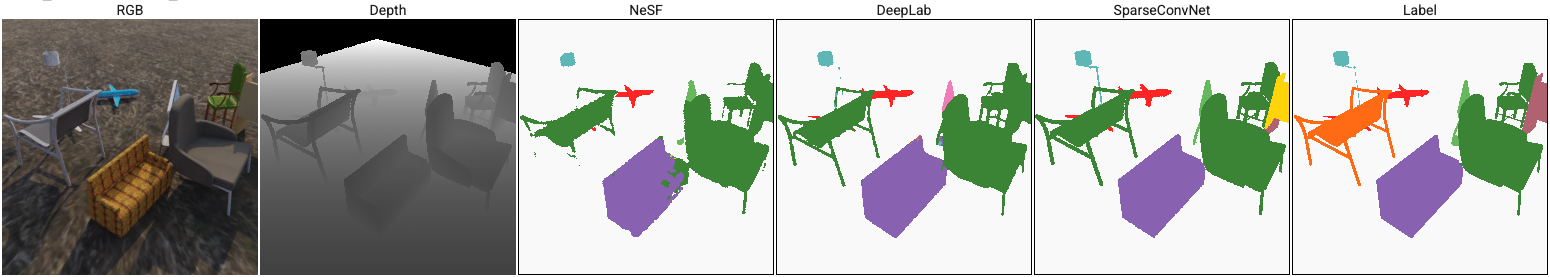}
    \includegraphics[width=0.95\linewidth]{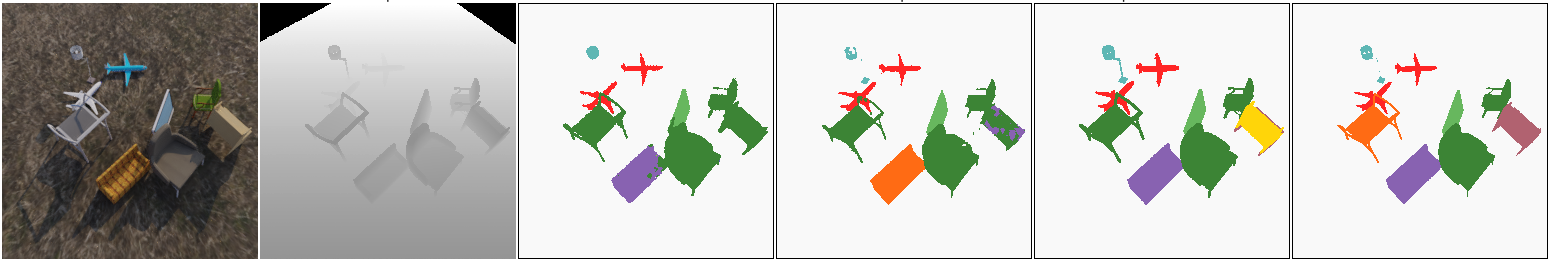}
\vspace{-1em}
\caption{
    \textbf{Multiview Consistency}
    While \NeSF{} and \SparseConvNet{} classify the orange couch and the display identically from multiple independent views of the same scene, \DeepLab{}'s predictions vary.
}
\label{fig:multiview_consistency}
\end{figure*}

\begin{figure*}
\centering
\includegraphics[width=0.75\textwidth,keepaspectratio]{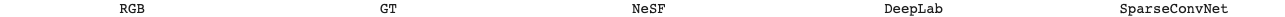}
\includegraphics[width=0.75\textwidth,keepaspectratio]{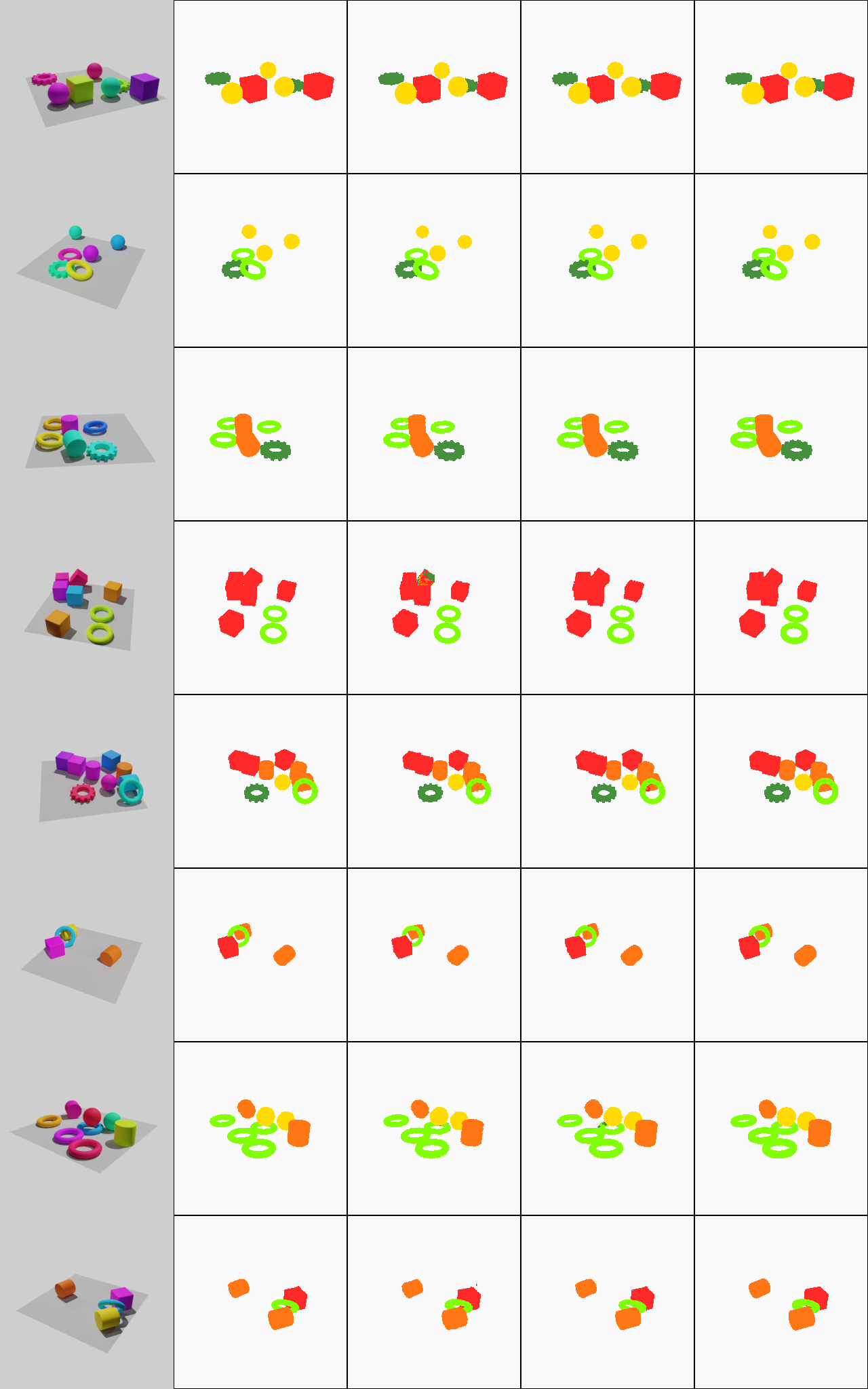}
\caption{
    \textbf{Additional Qualitative Results on \KLEVR{}}
}
\label{fig:qualitative_supplement_klevr}
\end{figure*}

\begin{figure*}
\centering
\includegraphics[width=0.75\textwidth,keepaspectratio]{fig/qualitative_supplement/header.png}
\includegraphics[width=0.75\textwidth,keepaspectratio]{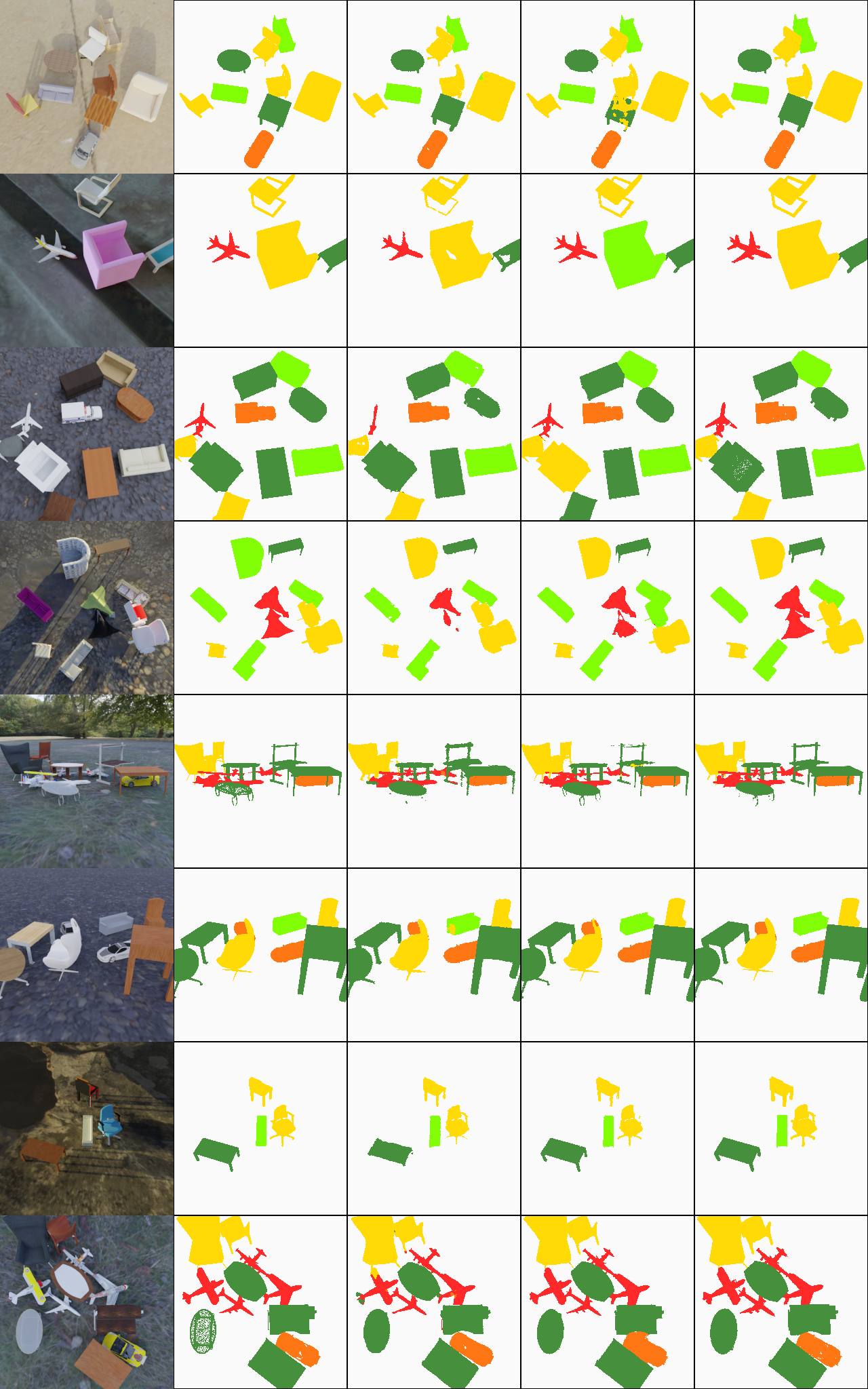}
\caption{
    \textbf{Additional Qualitative Results on \ToyBoxEasy{}}
}
\label{fig:qualitative_supplement_multi_shapenet_v05h5}
\end{figure*}

\begin{figure*}
\centering
\includegraphics[width=0.75\textwidth,keepaspectratio]{fig/qualitative_supplement/header.png}
\includegraphics[width=0.75\textwidth,keepaspectratio]{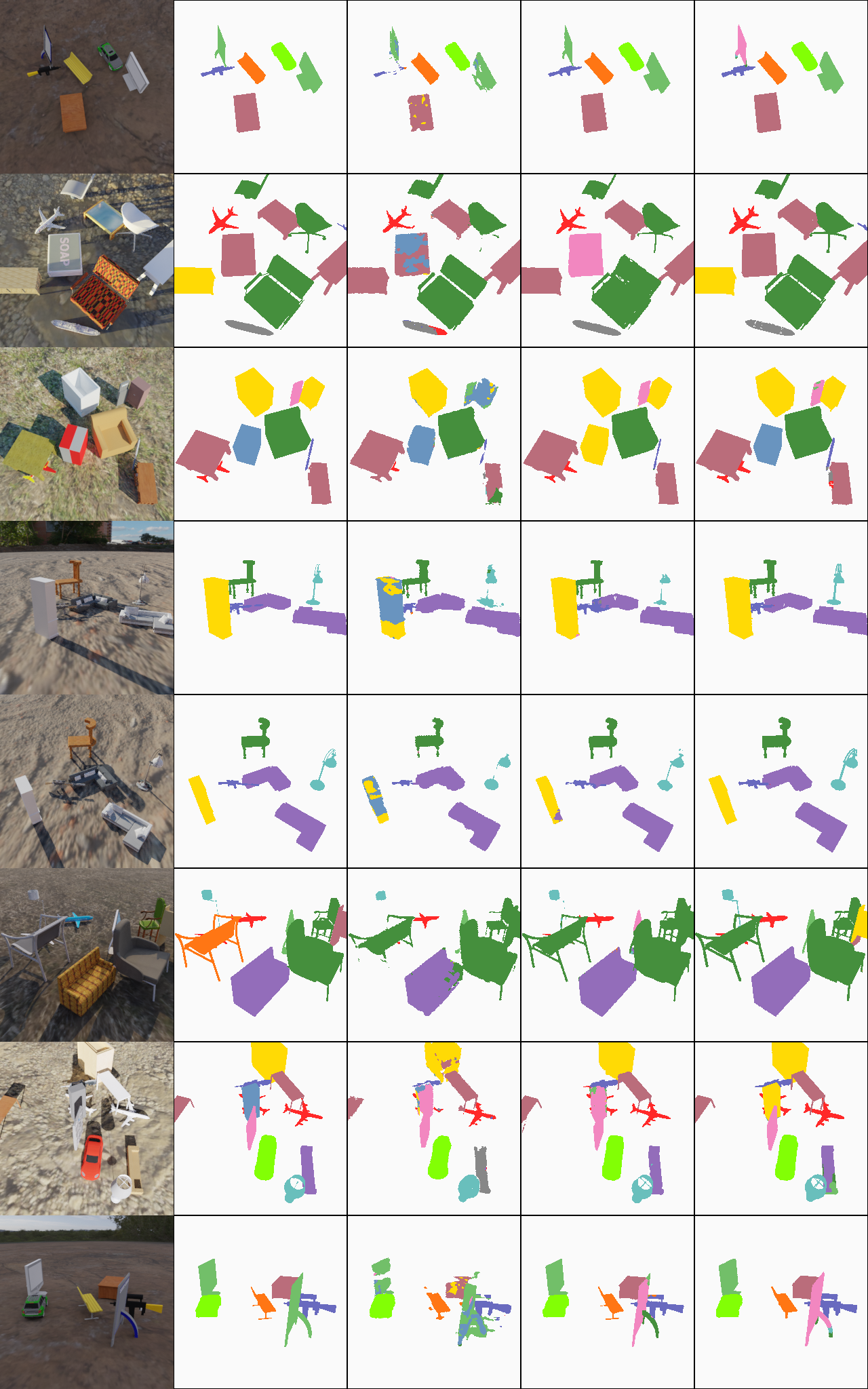}
\caption{
    \textbf{Additional Qualitative Results on \ToyBoxHard{}}
}
\label{fig:qualitative_supplement_multi_shapenet_v05h13}
\end{figure*}

In Figures~\ref{fig:qualitative_supplement_klevr}, \ref{fig:qualitative_supplement_multi_shapenet_v05h5}, and \ref{fig:qualitative_supplement_multi_shapenet_v05h13}, we present randomly-selected qualitative results on each dataset studied in this paper.
In each row, we depict the ground truth RGB, depth, and semantic map alongside 2D segmentation maps produced by \NeSF{}, \DeepLab{}, and \SparseConvNet{}.
We observe that all methods are effective at separating foreground objects from the floor and background.
Unlike \SparseConvNet{}, \NeSF{} and \DeepLab{} tend to assign different parts of the same object to different semantic categories when the correct category is ambiguous.

While \NeSF{} and \SparseConvNet{} are multi-view consistent by design, this is not the case for 2D methods such as \DeepLab{}.
In \Figure{multiview_consistency}, we demonstrate one instance of 3D inconsistency.
In this example, \NeSF{} and \SparseConvNet label the orange couch and white-blue display identically from both views, whereas \DeepLab{}'s classification changes.

\begin{figure*}
\centering
\includegraphics[width=0.75\textwidth,keepaspectratio]{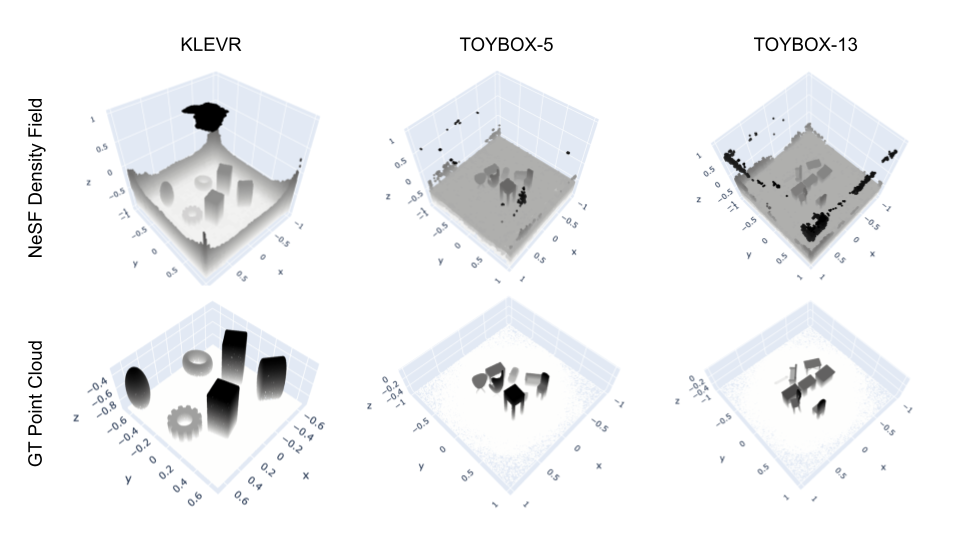}
\caption{
    \textbf{\NeRF{} 3D Density Field (top) and Ground Truth Point Clouds (bottom) for \KLEVR{}, \ToyBoxEasy{}, and \ToyBoxHard{}.}
    We find that \NeRF{}'s density field accurately captures the 3D geometry of the scene. The \NeSF{} density fields are sampled at a resolution of 128x128x128, and are filtered for ease of visualization for positive $\density$ values with thresholds of 16, 64, and 64 for the 3 datasets respectively.
}
\label{fig:qualitative_supplement_3d}
\end{figure*} 
\paragraph{\NeSF{} 3D Density Field Quality}
Notably, a key difference between \SparseConvNet and \NeSF{} is the provision of a ground truth point cloud as input for \SparseConvNet. Several aspects of \SparseConvNet may contribute to its overall superior performance relative to \NeSF including access to oracle 3D geometry, sparse point cloud input representation, or the \SparseConvNet model architecture. 
To better understand where headroom exists for improvement of \NeSF, we begin by visually inspecting the difference in 3D geometry between the 3D density field of \NeSF in relation to the ground truth point cloud provided to \SparseConvNet for scenes selected from \KLEVR{}, \ToyBoxEasy{}, and \ToyBoxHard{} in \Figure{qualitative_supplement_3d}. 
We observe the \NeSF density fields often miss thin structures and fine details, and ``floaters'' are particularly evident in \ToyBoxHard{}. 
Improvement of density field quality via improvements to \NeRF{} representations may resolve such artifacts. 
Furthermore, akin to results in Table~\ref{tab:reconstruction_quality_supp}, such improvements would likely improve the performance of \NeSF{}.
We leave additional inspection of 3D input representation and replacement of the semantic model architecture of \NeSF{} for future work. 

\subsection{Ablations}

\begin{table}
\small
\centering
\begin{tabular}{c|c|c|c} 
\toprule
\multicolumn{2}{c|}{Hyperparameter}
  & 2D & 3D \\
\hline
\multirow{2}{*}{\shortstack{Random \\ Rotations}}
  & No
  & 69.5  & 83.6 \\
  & Yes
  & 78.8  & 89.7 \\
\hline

\multirow{3}{*}{\shortstack{Density \\ Grid}}
  & (32, 32, 32)
  & 71.1  & 81.5 \\
  & (48, 48, 48)
  & 76.4  & 89.3 \\
  & (64, 64, 64)
  & 78.8  & 89.7 \\
\hline

\multirow{3}{*}{UNet}
  & (16, 32, 64)
  & 80.6  & 89.1 \\
  & (24, 48, 96)
  & 80.1  & 89.8 \\
  & (32, 64, 128)
  & 79.0  & 89.8 \\
\hline

\multirow{4}{*}{MLP}
  & (0, 32)
  & 78.6  & 90.7 \\
  & (1, 32)
  & 79.7  & 89.8 \\
  & (1, 64)
  & 80.7  & 89.4 \\
  & (2, 128)
  & 79.2  & 89.5 \\

\bottomrule
\end{tabular}
\caption{
\textbf{Ablation: hyper-parameters} --
Data augmentation, in the form of random scene rotations, increased spatial resolution of the density grid, and increased UNet model capacity improve 2D and 3D \mIoU{}.
Experiments on 500 scenes from the \ToyBoxEasy{} dataset.
}
\label{tab:nesf_ablations_supp}
\end{table}

\begin{table}
\small
\centering
\begin{tabular}{ccc|cc}
\toprule
\multirow{2}{*}{\shortstack{\# RGB \\ Images}}
  & \multicolumn{2}{c|}{\NeRF{}}
  & \multicolumn{2}{c}{\NeSF{}} \\
& PSNR & SSIM
  & 2D & 3D \\
\hline
5    & $17.5 \pm 2.1$   & $0.55 \pm 0.15$   & $15.0$   & $17.9$  \\
10   & $19.2 \pm 2.9$   & $0.62 \pm 0.15$   & $29.1$   & $35.2$  \\
25   & $23.9 \pm 2.7$   & $0.76 \pm 0.09$   & $61.4$   & $74.1$  \\
50   & $26.3 \pm 2.1$   & $0.81 \pm 0.06$   & $72.3$   & $88.7$  \\
75   & $27.3 \pm 2.0$   & $0.83 \pm 0.05$   & $72.6$   & $89.5$  \\
100  & $27.9 \pm 2.0$   & $0.84 \pm 0.04$   & $73.6$   & $90.0$  \\
\bottomrule
\end{tabular}
\caption{
\textbf{Ablation: sensitivity to reconstruction quality for \ToyBoxEasy} --
The accuracy of our method improves with \NeRF{}'s reconstruction quality.
PSNR and SSIM are are averaged across all scenes and metrics aggregated. Experiments on all scenes from \ToyBoxEasy{}.
}
\label{tab:reconstruction_quality_supp}
\end{table}

% Figure generated by Colab,
% https://colab.corp.google.com/drive/1vbeqGF63xHScQzzeHi4c5LUGWURV8bQ2?resourcekey=0-j_6KEX0t08UWOryMF-IHag#scrollTo=5xkUL1E7ijs2&line=26&uniqifier=1

\begin{figure}[t]
\centering
\includegraphics[width=0.44\linewidth]{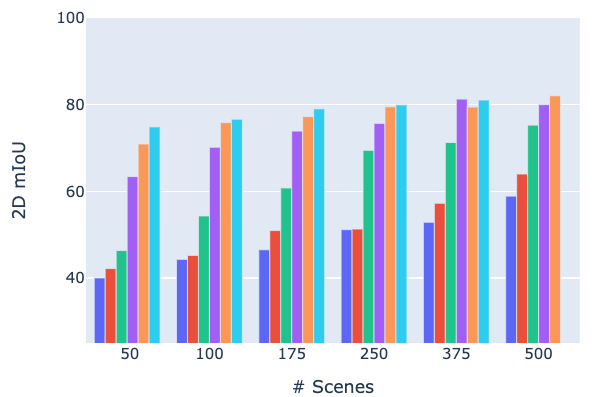}
\includegraphics[width=0.43\linewidth]{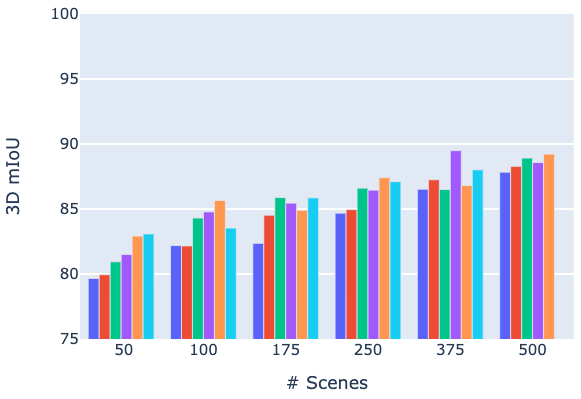}
\vspace{-.5em}
\caption{
\textbf{Ablation: data efficiency} --
2D and 3D mIoU as a function of the number of train scenes for scenes with supervision from~{\color{blue} \textbf{1}},~{\color{red} \textbf{2}},~{\color{ForestGreen} \textbf{5}},~{\color{purple} \textbf{10}}, ~{\color{YellowOrange} \textbf{25}}, or {\color{cyan} \textbf{50}}~semantic maps per scene. Note, {\color{cyan} \textbf{50}} semantic maps did not fit in memory for 500 scenes, hence this particular setup is dropped from the results.
\NeSF{} generalizes to new scenes with as few as a \emph{one semantic map per scene}.
Additional semantic maps per scene marginally improve the accuracy.
Experiments on \ToyBoxEasy{} dataset.
}
\label{fig:num_images_per_scene_supp}
\end{figure}

\paragraph{Model Ablations -- \Table{nesf_ablations_supp}}
We repeat our our ablation study on the \ToyBoxEasy{} model and observe overall consistent results with \KLEVR{} model ablations. \Table{nesf_ablations_supp} shows results varying each component. Similar to the results on \KLEVR{}, we observe that data augmentation in the form of random scene rotations improves quality the most, adding 9.3\% and 6.1\% to 2D and 3D \mIoU{} respectively.
The spatial resolution of the probed \NeRF{} density grids is again confirmed as crucial, and notably to a greater extent than for \KLEVR{}. We hypothesize that this occurs as \ToyBoxEasy{} contains more fine structured objects than \KLEVR{}. 

\paragraph{Sensitivity to reconstruction quality -- \Table{reconstruction_quality_supp}}
We re-evaluate the robustness of \NeSF{} to \NeRF{} reconstruction quality in the context of the \ToyBoxEasy{} dataset.
To modulate reconstruction quality, we vary the number of RGB images used when fitting \NeRF{} models from 5 to 100 and confirm \NeRF{} reconstruction quality improves as more RGB images are provided. As previously observed, the 2D and 3D segmentation quality of \NeSF{} improves monotonically with \NeRF{}'s reconstruction quality. Notably, 3D segmentation accuracy begins to level off near 88\% when \NeRF{} models are optimized with as few as 50 RGB images, with a large jump in performance between 25 and 50 images.

\boldpnov{Sensitivity to data scarcity -- \Figure{num_images_per_scene_supp}}
We repeat our analysis providing limited numbers of semantically labelled maps for \NeSF's training on the \ToyBoxEasy{} model. We vary the number of provided label maps from 1 to 50. Similar to the \KLEVR{} setup, we observe that providing additional semantic maps per scene improves the performance, with a large jump between 5 and 10 maps. A saturation in the model performance is reached at around 25 maps per scene. Moreover, the model is still able to generalize with as little as 1 semantic map per scene.

\TODO{Why we couldn't use COLMAP point clouds for an ablation? Justify here.}

\subsection{Multivew Consistency}

% Figure from this slide,
% https://colab.corp.google.com/drive/1Ac4oCld4E8uCAJYDpw4qgdZpeRCU3qGV?resourcekey=0-na0gkrIDeli7yDUv1EILqw#scrollTo=vcbf8Ur-zQa7&line=18&uniqifier=1

\begin{figure*}[t]
\centering
\includegraphics[width=0.9\linewidth]{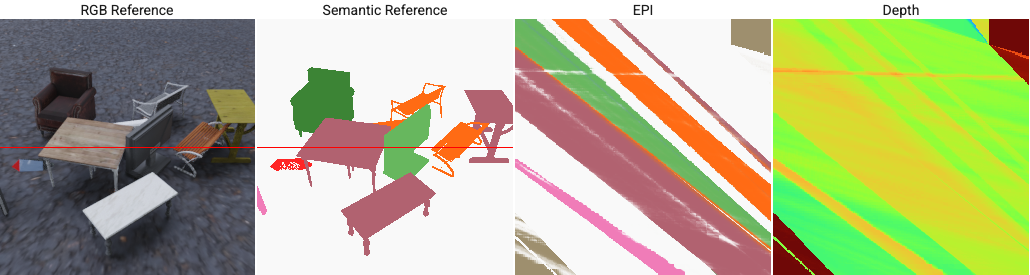}
\caption{
    \textbf{Epipolar Plane}
    We demonstrate the 3D consistency of \NeSF{} by rendering the epipolar plan along the red scan line as the camera moves from right to left.
    The epipolar plane is is smooth and consistent except when a ``floater'' passes in front of the camera.
}
\label{fig:epipolar}
\end{figure*}
% Figure from this slide,
% https://colab.corp.google.com/drive/1Ac4oCld4E8uCAJYDpw4qgdZpeRCU3qGV?resourcekey=0-na0gkrIDeli7yDUv1EILqw#scrollTo=4lCxBIE35mJf&line=3&uniqifier=1

\begin{figure*}[t]
\centering
    \includegraphics[width=0.90\linewidth]{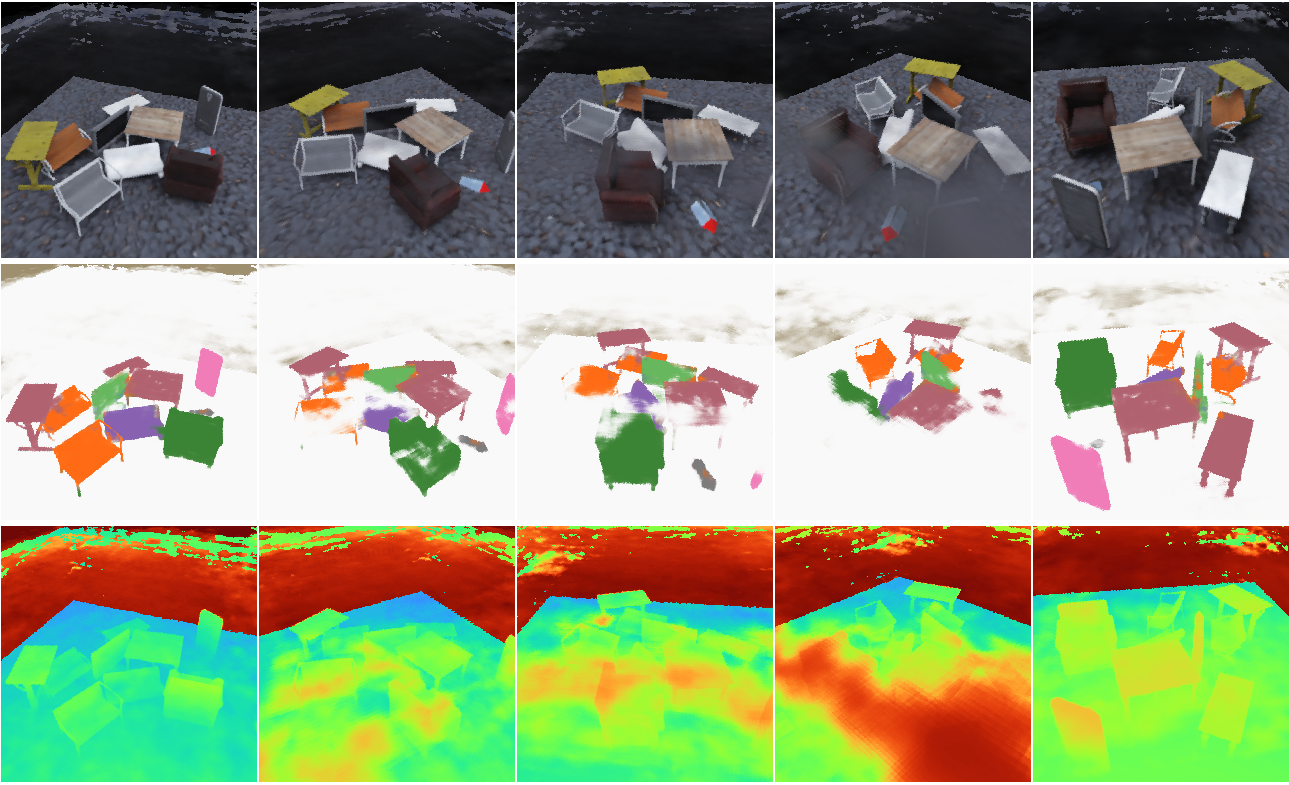}
\caption{
    \textbf{Floaters}
    In the above 3 rows, we illustrate \NeRF{}'s RGB reconstruction (top), \NeSF{}'s semantic field, and \NeRF{}'s density field (bottom) over the course of 5 video frames.
    When \NeRF{}'s density field contains ``floaters'', \NeSF{} often assigns them to the background semantic category.
}
\label{fig:floaters}
\end{figure*}

Unlike conventional 2D methods, \NeSF{} is 3D-consistent by design.
In \Figure{epipolar}, we visualize the epipolar plane traced out along the red reference line for \NeSF{}'s semantic predictions.
We find that the resulting predictions are consistent and smooth except when a floater obstructs the camera's view, as illustrated by the white smudge in the EPI.
In \Figure{floaters}, we see a further example of this phenomena.
As the camera rotates about the scene, a floating mass of density obstructs the camera's view, and the resulting semantic maps contain a large number of mislabeled pixels.
\WatchTheVideo{}

\subsection{Limitations}

\paragraph{Confusion matrix}
In \Table{confusion_matrix_2d} and \Table{confusion_matrix_3d}, we present \NeSF{}'s per-class confusion matrix on the \ToyBoxHard{} dataset for 2D pixel and 3D point classification.
While \NeSF{} is able to easily identify larger, articulated semantic categories such as cabinet, chair, display, or table (78.0-89.4\% 2D, 79.4-93.6\% 3D), it struggles to identify object categories for smaller objects such as rifle (56.3\% 2D, 75.3\% 3D) or geometrically unarticulated objects such as loudspeaker (38.5\% 2D, 40.4\% 3D).
When \NeSF{} confuses foreground object categories, the most common errors are between geometrically-similar classes.
For example, benches are often mislabeled as chairs (17.0\% 2D, 17.9\% 3D) and sofas (26.5\% 2D, 27.4\% 3D), and loudspeakers are often mislabeled as tables (32.0\% 2D, 32.5\% 3D).

\paragraph{Accuracy 2D~vs.~3D}
Our experiments indicate that \NeSF{}'s accuracy is \emph{higher} in 3D than in 2D.
We found this surprising, especially considering the 2D nature of \NeSF{}'s semantic supervision.
We believe the ultimate cause to be ``floaters'' in the 3D density field recovered by \NeRF{}.
In \Table{confusion_matrix_2d}, we see that approximately 5-10\% of 2D pixels from each semantic category are mislabeled as ``background''.
In contrast, \Table{confusion_matrix_3d} demonstrates that the same type of error is made approximately 1\% of the time in 3D.
The most prominent exception to this is the bench category, whose objects often contain thin structures poorly captured by \NeSF{}.

\paragraph{Impact of floaters}
In \Figure{floaters}, we qualitatively show how ``floaters'' reduce \NeSF{}'s accuracy in image-space. 
In this set of 5 video frames, we demonstrate a camera path passing in front of a floating cloud.
This cloud is assigned to the background semantic category and obscures the foreground objects from the scene.
In spite of the \emph{objects} being correctly labeled, the generated semantic maps are incorrect.
As a result, \NeSF{} achieves lower 2D \mIoU{} than 3D \mIoU{} as the latter is not hindered by floaters and is corroborated by Tables~\ref{tab:confusion_matrix_2d} and~\ref{tab:confusion_matrix_3d}.
We believe that eliminating this failure in geometric construction will significantly improve \NeSF{}'s accuracy.
Solutions are readily provided by methods building on \NeRF{}~\cite{Yu21iccv_PlenOctrees}.
\WatchTheVideo{}

\begin{landscape}

\begin{table*}[ht]
\tiny
\centering
\begin{tabular}{c|cccccccccccccc} 
\toprule
            & background & airplane &  bench & cabinet &    car &  chair & display &   lamp & loudspeaker &  rifle &   sofa &  table & telephone & vessel \\
\hline
background	&	\textbf{99.1\%}	&	 0.1\%	&	 0.0\%	&	 0.0\%	&	 0.0\%	&	 0.2\%	&	 0.1\%	&	 0.0\%	&	 0.0\%	&	 0.0\%	&	 0.0\%	&	 0.2\%	&	 0.0\%	&	 0.0\%	\\
airplane	&	12.9\%	&	\textbf{69.2\%}	&	 3.2\%	&	 0.1\%	&	 0.0\%	&	 1.9\%	&	 0.0\%	&	 0.3\%	&	 0.0\%	&	 1.4\%	&	 0.0\%	&	 4.6\%	&	 0.0\%	&	 6.3\%	\\
bench	&	 9.1\%	&	 0.7\%	&	\textbf{42.1\%}	&	 0.1\%	&	 0.0\%	&	17.0\%	&	 0.0\%	&	 0.2\%	&	 0.0\%	&	 1.2\%	&	26.5\%	&	 1.7\%	&	 0.0\%	&	 1.4\%	\\
cabinet	&	 2.2\%	&	 0.0\%	&	 0.0\%	&	\textbf{78.0\%}	&	 0.0\%	&	 0.1\%	&	 2.7\%	&	 0.0\%	&	12.8\%	&	 0.0\%	&	 0.0\%	&	 2.7\%	&	 1.5\%	&	 0.0\%	\\
car	&	 5.0\%	&	 0.1\%	&	 0.1\%	&	 0.0\%	&	\textbf{84.6\%}	&	 0.5\%	&	 0.1\%	&	 0.3\%	&	 0.3\%	&	 0.2\%	&	 1.3\%	&	 4.1\%	&	 0.0\%	&	 3.3\%	\\
chair	&	 5.8\%	&	 0.0\%	&	 0.9\%	&	 0.0\%	&	 0.0\%	&	\textbf{89.4\%}	&	 0.1\%	&	 0.0\%	&	 0.1\%	&	 0.0\%	&	 2.9\%	&	 0.7\%	&	 0.0\%	&	 0.0\%	\\
display	&	 9.2\%	&	 0.0\%	&	 0.3\%	&	 0.4\%	&	 0.1\%	&	 1.2\%	&	\textbf{83.3\%}	&	 0.0\%	&	 3.7\%	&	 0.0\%	&	 0.2\%	&	 1.3\%	&	 0.2\%	&	 0.0\%	\\
lamp	&	11.1\%	&	 0.0\%	&	 0.0\%	&	 0.0\%	&	 0.0\%	&	 0.5\%	&	 0.0\%	&	\textbf{61.4\%}	&	18.1\%	&	 0.0\%	&	 0.0\%	&	 0.1\%	&	 7.8\%	&	 0.8\%	\\
loudspeake	&	 5.4\%	&	 0.1\%	&	 4.3\%	&	10.5\%	&	 0.0\%	&	 0.6\%	&	 3.2\%	&	 1.1\%	&	\textbf{38.5\%}	&	 0.0\%	&	 2.2\%	&	32.0\%	&	 1.7\%	&	 0.4\%	\\
rifle	&	30.7\%	&	 1.6\%	&	 2.0\%	&	 0.2\%	&	 0.0\%	&	 0.7\%	&	 3.1\%	&	 0.0\%	&	 0.0\%	&	\textbf{56.3\%}	&	 3.0\%	&	 0.7\%	&	 0.0\%	&	 1.7\%	\\
sofa	&	10.0\%	&	 2.3\%	&	 0.6\%	&	 0.0\%	&	 0.0\%	&	21.6\%	&	 0.0\%	&	 0.1\%	&	 0.1\%	&	 0.0\%	&	\textbf{61.6\%}	&	 0.0\%	&	 0.0\%	&	 3.6\%	\\
table	&	 5.3\%	&	 0.0\%	&	 1.7\%	&	 3.1\%	&	 0.0\%	&	 4.6\%	&	 0.1\%	&	 3.1\%	&	 1.2\%	&	 0.0\%	&	 0.7\%	&	\textbf{79.9\%}	&	 0.0\%	&	 0.3\%	\\
telephone	&	 1.7\%	&	 0.0\%	&	 0.0\%	&	 9.3\%	&	 0.0\%	&	 0.2\%	&	 3.7\%	&	 0.0\%	&	15.7\%	&	 0.0\%	&	 0.0\%	&	 0.3\%	&	\textbf{69.0\%}	&	 0.0\%	\\
vessel	&	 5.9\%	&	12.4\%	&	 0.0\%	&	 0.0\%	&	 0.0\%	&	 0.1\%	&	 0.0\%	&	 0.0\%	&	 0.0\%	&	 4.3\%	&	 0.3\%	&	 0.0\%	&	 0.0\%	&	\textbf{77.0\%}	\\
\bottomrule
\end{tabular}
\caption{
    \textbf{Confusion matrix for 2D semantic segmentations by \NeSF{} on \ToyBoxHard{}.}
    Each row corresponds to a ground truth label and is normalized to sum to 100\%.
    \NeSF{}'s most common errors include confusing similarly-shaped objects and classifying small and thin objects as background.
    Correct classifications are highlighted in bold.
}
\label{tab:confusion_matrix_2d}
\end{table*}

\begin{table*}[ht]
\tiny
\centering
\begin{tabular}{c|cccccccccccccc} 
\toprule
            & background & airplane &  bench & cabinet &    car &  chair & display &   lamp & loudspeaker &  rifle &   sofa &  table & telephone & vessel \\
\hline
background	&	\textbf{99.4\%}	&	 0.0\%	&	 0.0\%	&	 0.0\%	&	 0.0\%	&	 0.0\%	&	 0.1\%	&	 0.1\%	&	 0.0\%	&	 0.0\%	&	 0.0\%	&	 0.1\%	&	 0.0\%	&	 0.0\%	\\
airplane	&	 1.7\%	&	\textbf{78.0\%}	&	 3.6\%	&	 0.0\%	&	 0.0\%	&	 1.8\%	&	 0.0\%	&	 0.2\%	&	 0.0\%	&	 2.1\%	&	 0.0\%	&	 4.8\%	&	 0.1\%	&	 7.6\%	\\
bench	&	 6.0\%	&	 0.8\%	&	\textbf{42.7\%}	&	 0.2\%	&	 0.0\%	&	17.9\%	&	 0.0\%	&	 0.3\%	&	 0.0\%	&	 1.7\%	&	27.4\%	&	 1.6\%	&	 0.0\%	&	 1.4\%	\\
cabinet	&	 0.4\%	&	 0.0\%	&	 0.0\%	&	\textbf{79.4\%}	&	 0.0\%	&	 0.0\%	&	 2.9\%	&	 0.0\%	&	13.0\%	&	 0.0\%	&	 0.0\%	&	 2.6\%	&	 1.7\%	&	 0.0\%	\\
car	&	 0.8\%	&	 0.4\%	&	 0.3\%	&	 0.0\%	&	\textbf{86.7\%}	&	 0.6\%	&	 0.0\%	&	 0.2\%	&	 0.4\%	&	 0.3\%	&	 2.1\%	&	 4.0\%	&	 0.0\%	&	 4.3\%	\\
chair	&	 1.0\%	&	 0.0\%	&	 1.1\%	&	 0.0\%	&	 0.0\%	&	\textbf{93.6\%}	&	 0.1\%	&	 0.1\%	&	 0.1\%	&	 0.1\%	&	 3.1\%	&	 0.8\%	&	 0.0\%	&	 0.0\%	\\
display	&	 0.6\%	&	 0.1\%	&	 0.4\%	&	 0.3\%	&	 0.0\%	&	 1.0\%	&	\textbf{90.7\%}	&	 0.0\%	&	 4.8\%	&	 0.0\%	&	 0.4\%	&	 1.4\%	&	 0.3\%	&	 0.0\%	\\
lamp	&	 2.3\%	&	 0.0\%	&	 0.0\%	&	 0.1\%	&	 0.1\%	&	 0.4\%	&	 0.0\%	&	\textbf{66.7\%}	&	20.7\%	&	 0.0\%	&	 0.0\%	&	 0.1\%	&	 8.8\%	&	 0.9\%	\\
loudspeake	&	 1.1\%	&	 0.1\%	&	 4.4\%	&	12.5\%	&	 0.1\%	&	 0.4\%	&	 3.0\%	&	 1.2\%	&	\textbf{40.4\%}	&	 0.0\%	&	 2.1\%	&	32.5\%	&	 1.9\%	&	 0.4\%	\\
rifle	&	 2.4\%	&	 4.9\%	&	 4.6\%	&	 0.6\%	&	 0.1\%	&	 1.8\%	&	 3.6\%	&	 0.0\%	&	 0.0\%	&	\textbf{75.3\%}	&	 3.2\%	&	 0.9\%	&	 0.0\%	&	 2.7\%	\\
sofa	&	 0.7\%	&	 2.9\%	&	 0.8\%	&	 0.0\%	&	 0.0\%	&	22.8\%	&	 0.0\%	&	 0.1\%	&	 0.1\%	&	 0.0\%	&	\textbf{68.2\%}	&	 0.0\%	&	 0.0\%	&	 4.3\%	\\
table	&	 0.6\%	&	 0.0\%	&	 1.8\%	&	 3.5\%	&	 0.0\%	&	 4.7\%	&	 0.0\%	&	 3.2\%	&	 1.5\%	&	 0.1\%	&	 0.8\%	&	\textbf{82.8\%}	&	 0.0\%	&	 0.8\%	\\
telephone	&	 0.1\%	&	 0.0\%	&	 0.0\%	&	10.5\%	&	 0.0\%	&	 0.0\%	&	 4.5\%	&	 0.0\%	&	15.8\%	&	 0.0\%	&	 0.0\%	&	 0.0\%	&	\textbf{69.1\%}	&	 0.0\%	\\
vessel	&	 1.1\%	&	13.2\%	&	 0.0\%	&	 0.0\%	&	 0.0\%	&	 0.0\%	&	 0.0\%	&	 0.0\%	&	 0.0\%	&	 4.4\%	&	 0.6\%	&	 0.0\%	&	 0.0\%	&	\textbf{80.7\%}	\\
\bottomrule
\end{tabular}
\caption{
    \textbf{Confusion matrix for 3D semantic segmentations by \NeSF{} on \ToyBoxHard{}.}
    Each row corresponds to a ground truth label and is normalized to sum to 100\%.
    \NeSF{}'s most common errors include confusing similarly-shaped objects and classifying small and thin objects as background.
    Correct classifications are highlighted in bold.
}
\label{tab:confusion_matrix_3d}
\end{table*}
\end{landscape}


\end{document}